\documentclass{article}
\usepackage{preprint,times}
\usepackage{amsmath,amsfonts,bm}

\def\eqref#1{equation~\ref{#1}}
\def\1{\bm{1}}

\DeclareMathAlphabet{\mathsfit}{\encodingdefault}{\sfdefault}{m}{sl}
\SetMathAlphabet{\mathsfit}{bold}{\encodingdefault}{\sfdefault}{bx}{n}

\usepackage{booktabs}
\usepackage{tabularx}
\usepackage{float}
\usepackage{flafter}
\usepackage{graphicx}
\usepackage{wrapfig}
\usepackage{amsmath}
\usepackage{amssymb}
\usepackage{xspace}
\usepackage{enumitem}
\usepackage[table]{xcolor}
\usepackage{tikz}
\usepackage{etoc}
\usepackage{hyperref}
\hypersetup{hidelinks}
\usepackage{url}

\newcommand{\methodn}{GSO}
\newcommand{\method}{\textsc{\methodn}\xspace}

\title{Do Self-Evolving Skills Generalize to \\Held-Out Tasks?}

\author{Xihao Piao$^{1}$ \quad Zifeng Wang$^{\dagger}$ \quad Zheng Chen$^{1}$ \\
$^{1}$SANKEN, Osaka University \\
\texttt{park88@sanken.osaka-u.ac.jp} \\
$^{\dagger}$Corresponding author}

\ppfinalcopy

\begin{document}
\maketitle
\fancyhead{}\renewcommand{\headrulewidth}{0pt}

\begin{abstract}
AI agents can externalize what they learn from past tasks into reusable \emph{skills}, such as procedures, checklists, code, or other executable artifacts, that can be retrieved and reused when solving new tasks.
Self-evolving skill methods keep rewriting these skills after each round of practice on training tasks, and the skill is then used on new tasks of the same kind.
We ask a question: does the improvement a skill shows on its training tasks carry over to new test tasks?
We test five self-evolving methods and a one-shot skill on six benchmarks, with the same model, the same agent, and the same train/test split for every method.
Of the 21 skills that improve on their training tasks, 5 keep all of that improvement on the test tasks, 13 keep part of it, and 3 keep none of it.
No existing method is best everywhere.
When we read the skills, the ones that carry over badly often fix details that should depend on the task, such as column names and output files, or turn a fix for one failure into a rule for every task.
An LLM judge that reads the skill content can often see this: it ranks finished skills the same way the test results do in 86\% of pairs.
But it predicts the effect of a single edit poorly, so edits still have to be tested by running them.
Based on these findings, we describe Generalizable Skill Optimization (\method), which keeps only a guide for writing skills and writes a new skill for each task; it scores highest on all six benchmarks.
\end{abstract}

\section{Introduction}

\begin{figure*}[t]
    \centering
    \includegraphics[width=\textwidth]{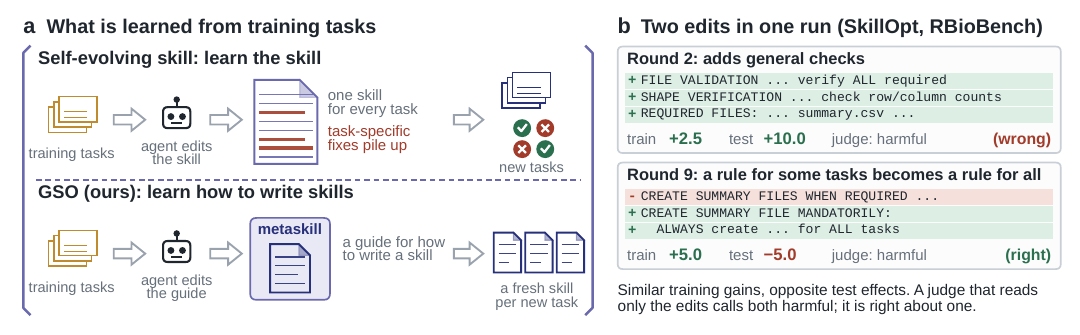}
    \caption{(a) Existing methods optimize skills across training tasks and reuse them on new tasks.
        We instead optimize a metaskill that generates a new skill for each task.
        (b) Two edits from the same SkillOpt run on RBioBench, quoted verbatim.
        Both improve training performance slightly, yet one helps on test tasks while the other hurts.
        An LLM judge that reads only the edit predicts that both will hurt.
        Scores are changes in success rate from the previous skill (percentage points).}
    \label{fig:story}
\end{figure*}

An AI agent often solves many tasks, such as editing spreadsheets or fixing bugs in code.
It would help if the agent could learn from the tasks it has already done.
A popular way to do this is to maintain a \emph{skill}: a reusable artifact that captures what the agent has learned, such as a procedure, checklist, prompt, or executable code, and can be applied when solving new tasks.
Self-evolving skill methods go one step further.
They let the agent practice on training tasks, use the resulting feedback, and iteratively revise the skill \citep{wang2023voyager,ma2026skillgen,ni2026trace2skill,yang2026skillopt,alzubi2026evoskill}.

\begin{figure*}[t]
    \centering
    \includegraphics[width=\textwidth]{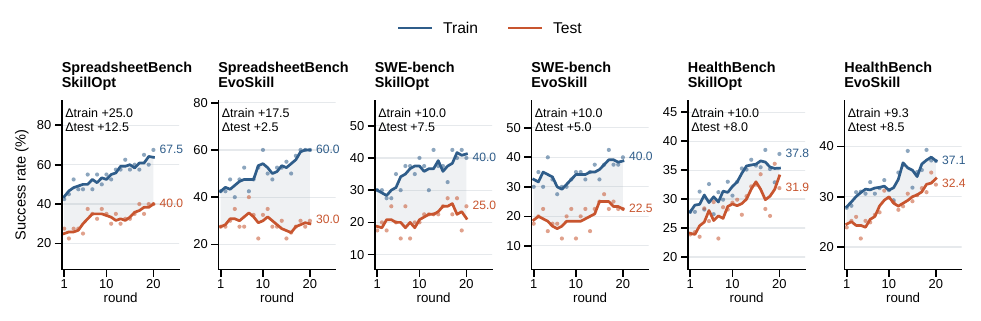}
    \caption{Training and test success rates of SkillOpt and EvoSkill over twenty rounds on three benchmarks.
        Dots are per-round values and lines a three-round moving average; $\Delta$ is the change from round 1 to round 20, and shading marks the gap between training and test.
        On SWE-bench and HealthBench the test gain follows the training gain; on SpreadsheetBench it lags far behind.}
    \label{fig:train-test-generalization-case}
\end{figure*}

The point of all this rewriting is to do better on new tasks, not only on the training tasks.
Figure~\ref{fig:story}b shows that this does not always happen.
It follows two edits that SkillOpt proposed in one run.
Both raise the training score a little.
On new test tasks the first raises the score by 10 points and the second lowers it by 5.
The second edit turned a rule for some tasks (``create a summary file when one is required'') into a rule for all tasks (``always create one'').
Across methods and benchmarks, this is common: of 21 final skills that improve on their training tasks, only 5 keep the whole improvement on the test tasks (Figure~\ref{fig:train-test-generalization-case}, Section~\ref{sec:generalization}).

Why would this happen?
In each round, the method changes the skill so that it passes the training tasks it just failed, and the changed skill is used on every future task.
A fix that works for one task may not work for another, and after many rounds the skill can turn into a list of such fixes.
As in other kinds of machine learning, we call this \emph{skill overfitting}.

This paper first defines and measures skill overfitting.
We evaluate six methods and a no-skill agent across six benchmarks in a controlled setting, using a common model, agent architecture, and train/validation/test split, while keeping test tasks hidden until each method selects its final skill.
We then inspect the learned skills to identify patterns associated with overfitting and test whether an LLM judge can detect them.
This analysis suggests a different learning target (Figure~\ref{fig:story}a): instead of learning one skill to reuse across tasks, learn how to write a skill and write a fresh one for each task.
We build \method on this idea and test it the same way.
Our contributions are:
\begin{itemize}[leftmargin=*, itemsep=1pt, topsep=2pt]
    \item A fair test of whether self-evolved skills carry over to new tasks of the same kind.
        Of the 21 skills from existing methods that improve on their training tasks, only 5 keep the whole improvement on the test tasks, and 10 of the 36 test scores of existing methods (six methods on six benchmarks) are lower than No Skill.
    \item An analysis of which skills carry over.
        Skills that carry over badly copy details of their training tasks into rules for every task (Figure~\ref{fig:skill-content}).
        A judge that reads only the skill agrees with the test results on 108 of 126 pairs of finished skills, but it predicts the effect of a single edit poorly.
    \item \method, built from this analysis.
        It keeps only a guide for writing skills and writes a fresh skill for each task.
        It is best on all six benchmarks, 4.3 to 22.5 points above the best existing method.
\end{itemize}

\section{Related Work}
\label{sec:related-work}

\paragraph{Memory and skills.} Agents can keep what they learn from past tasks in several forms.
Memory methods keep notes, such as a reflection on why an attempt failed \citep{shinn2023reflexion}, lessons drawn from many attempts \citep{zhao2024expel}, or reasoning strategies \citep{ouyang2026reasoningbank}.
Skill methods keep something the agent can follow or run, such as a library of code \citep{wang2023voyager}, a tool \citep{cai2024latm}, or a workflow \citep{wang2024awm}.
Self-evolving skill methods add a loop: after each round of practice, they rewrite the skill based on what went wrong.
SkillGen compares successful and failed attempts \citep{ma2026skillgen}, Trace2Skill merges lessons from many attempts \citep{ni2026trace2skill}, SkillOpt edits one document and keeps an edit only if a validation check passes \citep{yang2026skillopt}, and EvoSkill grows a folder of skills \citep{alzubi2026evoskill}; related systems keep such skills in a memory that grows with experience \citep{fang2026memp,yang2026muse}.
All of them assume that skills learned from past tasks will help on the next task of the same kind; few measure how much of the training gain is left on new tasks.

\paragraph{Prompt optimization and self-improvement.} A related line of work optimizes parts of an agent based on task feedback.
ProTeGi edits prompts in response to errors \citep{pryzant2023protegi}, DSPy tunes pipelines of prompts to improve an objective \citep{khattab2024dspy}, and GEPA maintains and revises multiple candidates based on feedback \citep{agrawal2025gepa}.
Other work lets a meta agent write new agents in code \citep{hu2024adas}, and recursive self-improvement lets agents modify their own code \citep{zelikman2024stop,zhang2025dgm}, while test-time learning continues adapting the system during deployment \citep{suzgun2026dynamic}.
These approaches differ in what is adapted, memory, skills, prompts, pipelines, or code, and in how much of the system is allowed to change.
Their commonality is that persistent components are updated from past feedback and can cause overfitting to the tasks used for adaptation.
Skills provide an especially interpretable setting for studying this problem because the learned artifact can be inspected directly: we can ask whether it captures a reusable procedure or merely accumulates fixes for previously seen tasks.
We use GEPA as a baseline and freeze learned skills before testing, so test-time adaptation is not mixed up with transfer from prior tasks.

\paragraph{Transfer of stored experience.} Some studies already show that stored experience can hurt.
Memories can carry errors forward, and experience reused on a task it does not fit can mislead the agent \citep{xiong2026memorymanagement,liang2026remember}.
LLM judges, which we use to read skills, can change their answer when the order of the inputs changes \citep{zheng2023mtbench,shi2024judging}, so our judge scores each skill on its own instead of comparing two side by side.
We did not find a study that compares several skill-evolution methods with the same model, the same agent, and the same train/validation/test split, with every skill frozen before testing.
This paper provides one.
Appendix~\ref{app:related-work} discusses more related work.

\section{Problem Definition and Analysis}
\label{sec:generalization}

\begin{figure*}[t]
    \centering
    \includegraphics[width=\textwidth]{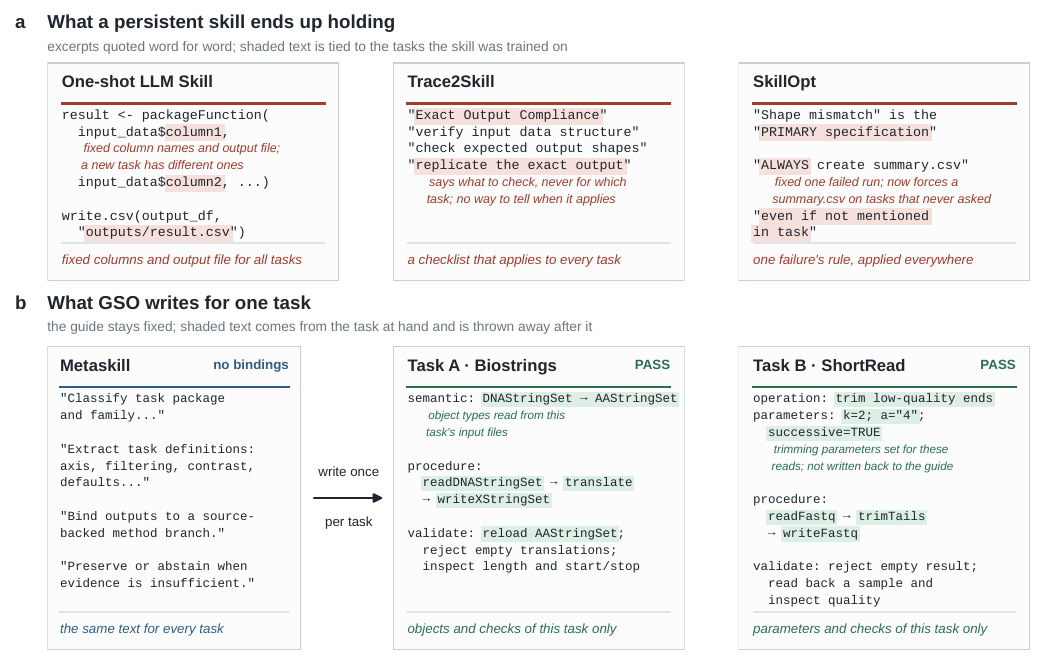}
    \caption{What an overfit skill looks like, and what \method writes instead.
        (a) Three skills learned by existing methods, quoted word for word.
        Each one fixes something that should depend on the task: column names and an output file, a checklist that applies to every task regardless of what the task asks, and a rule that every run must create a summary file.
        (b) Two skills that \method wrote for two tasks it had never seen.
        Both name the actual objects, functions, parameters, and checks of the task in front of them, and both passed.
        The metaskill on the left (\method's guide for writing skills, Section~\ref{sec:method}) stays the same; only the skill on the right changes with the task.}
    \label{fig:skill-content}
\end{figure*}

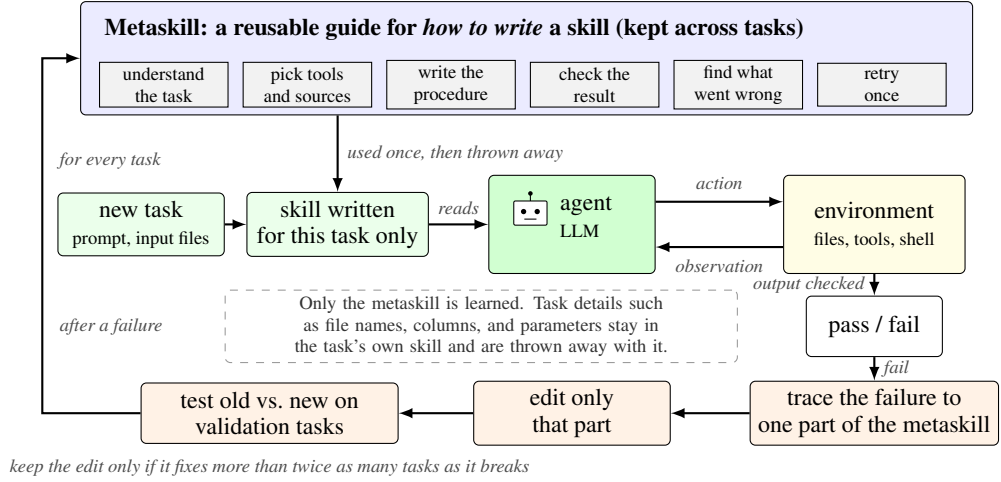
\begin{figure*}[t]
\centering
\begin{tikzpicture}[
  font=\small,
  box/.style={draw, rounded corners=2pt, align=center, inner sep=3pt, minimum height=8mm},
  mod/.style={draw, fill=gray!10, align=center, inner sep=2pt, font=\scriptsize, minimum width=17mm, minimum height=5.5mm},
  arr/.style={-latex, thick},
  lab/.style={font=\scriptsize\itshape, text=black!70}
]
% ---- metaskill (kept across tasks) ----
\node[box, fill=blue!8, minimum width=118mm, minimum height=15mm] (meta) at (0,0) {};
\node[anchor=north west, font=\small\bfseries] at ([xshift=2mm,yshift=-1mm]meta.north west) {Metaskill: a reusable guide for \emph{how to write} a skill (kept across tasks)};
\node[mod] at (-48mm,-3.5mm) {understand\\the task};
\node[mod] at (-29mm,-3.5mm) {pick tools\\and sources};
\node[mod] at (-10mm,-3.5mm) {write the\\procedure};
\node[mod] at (9mm,-3.5mm) {check the\\result};
\node[mod] at (28mm,-3.5mm) {find what\\went wrong};
\node[mod] at (47mm,-3.5mm) {retry\\once};
% ---- one task: agent loop ----
\node[lab, anchor=west] at (-63mm,-13.5mm) {for every task};
\node[box, fill=green!8, minimum width=22mm] (task) at (-51mm,-22mm) {new task\\\scriptsize prompt, input files};
\node[box, fill=green!8, minimum width=24mm] (skill) at (-25mm,-22mm) {skill written\\for this task only};
% agent box with a small robot icon
\node[box, fill=green!18, minimum width=22mm, minimum height=13mm] (agent) at (6mm,-22mm) {};
\begin{scope}[shift={(0.5mm,-20.5mm)}]
  \draw[rounded corners=0.6pt, fill=white] (-2.2mm,-1.6mm) rectangle (2.2mm,1.6mm);
  \fill (-0.9mm,0.2mm) circle (0.35mm); \fill (0.9mm,0.2mm) circle (0.35mm);
  \draw (-0.8mm,-0.8mm) -- (0.8mm,-0.8mm);
  \draw (0,1.6mm) -- (0,2.4mm); \fill (0,2.6mm) circle (0.3mm);
\end{scope}
\node[align=left, anchor=west, font=\small] at (3.2mm,-21mm) {agent\\\scriptsize LLM};
\node[box, fill=yellow!12, minimum width=24mm, minimum height=13mm, align=center] (env) at (46mm,-22mm) {environment\\\scriptsize files, tools, shell};
\draw[arr] (task) -- (skill);
\draw[arr] (skill) -- node[lab, above] {reads} (agent);
\draw[arr] ([yshift=3mm]agent.east) -- node[lab, above=0.3mm] {action} ([yshift=3mm]env.west);
\draw[arr] ([yshift=-3mm]env.west) -- node[lab, below=0.3mm] {observation} ([yshift=-3mm]agent.east);
\draw[arr] (meta.south -| skill) -- node[lab, right] {used once, then thrown away} (skill.north);
\node[box, minimum width=18mm] (out) at (46mm,-35.5mm) {pass / fail};
\draw[arr] (env.south) -- node[lab, left] {output checked} (out.north);
% ---- after a failure: update the metaskill ----
\node[lab, anchor=west] at (-63mm,-35.5mm) {after a failure};
\node[draw=black!40, dashed, rounded corners=2pt, align=center, font=\scriptsize, text=black!80, inner sep=2.5pt, text width=66mm] at (-6mm,-35.5mm) {Only the metaskill is learned. Task details such as file names, columns, and parameters stay in the task's own skill and are thrown away with it.};
\node[box, fill=orange!10, minimum width=30mm] (attr) at (46mm,-47mm) {trace the failure to\\one part of the metaskill};
\node[box, fill=orange!10, minimum width=26mm] (cand) at (6mm,-47mm) {edit only\\that part};
\node[box, fill=orange!10, minimum width=34mm] (gate) at (-34mm,-47mm) {test old vs.\ new on\\validation tasks};
\draw[arr] (out) -- node[lab, right] {fail} (attr);
\draw[arr] (attr) -- (cand);
\draw[arr] (cand) -- (gate);
\draw[arr] (gate.west) -- ++(-13mm,0) |- (meta.west);
\node[lab, anchor=north] at ([yshift=-0.8mm]gate.south) {keep the edit only if it fixes more than twice as many tasks as it breaks};
\end{tikzpicture}
\caption{How \method works.
    The only thing kept across tasks is a metaskill, a guide for how to write a skill.
    For each new task, a fresh skill is written from the task's own files and tools.
    The agent reads it and works in the environment in a loop, taking actions (such as running code) and reading what comes back, until the output is checked.
    The skill is then thrown away.
    When a task fails, the failure is traced to one part of the metaskill and only that part is edited; the edit stays only if it fixes more than twice as many validation tasks as it breaks.}
\label{fig:gso-framework}
\end{figure*}

\subsection{Definitions}

An agent solves a task by reading the task and acting in an environment.
Throughout this paper the agent, meaning the model and the program that runs its actions, is fixed.
What changes is a \emph{skill} $s$: a reusable artifact that the agent loads before it starts, such as a procedure, a checklist, a prompt, executable code, or a small folder of these.
We write $s_\varnothing$ for the empty skill, that is, the agent with no skill (No Skill).
For a task $x$, let $r(x,s)\in[0,1]$ be the score of the agent that reads $s$: 1 or 0 for pass or fail, or a graded score on benchmarks that give one.
Each benchmark contains tasks of one kind, such as spreadsheet edits or bug fixes, split into disjoint training, validation, and test sets $D_{\mathrm{train}}$, $D_{\mathrm{val}}$, and $D_{\mathrm{test}}$.

\textbf{Definition 1 (Self-evolving skill).} A self-evolving skill method starts from a first skill $s_0$ (often the empty skill $s_\varnothing$) and repeats two steps for $t=0,1,\dots,T-1$.
It runs the agent with $s_t$ on training tasks and collects feedback $F_t$, such as the runs that failed; then it writes a candidate $s'_t=E(s_t,F_t)$ with an editor $E$ and decides whether to keep it:
\begin{equation}
s_{t+1}=\begin{cases}s'_t & \text{if the candidate passes the method's check},\\ s_t & \text{otherwise.}\end{cases}
\label{eq:self-evolve}
\end{equation}
The check may use $D_{\mathrm{train}}$ or $D_{\mathrm{val}}$.
The final skill $\hat s$ (the last one, or one chosen on $D_{\mathrm{val}}$) is frozen and used on every test task.
Methods differ in the editor and the check (Appendix Table~\ref{tab:method-framework}).
Test tasks are never used to edit or choose a skill.

\textbf{Definition 2 (Generalization).} We study generalization across instances of the same task type.
A skill is learned from training instances of a task type and then evaluated on previously unseen instances of that same type.
For example, a spreadsheet skill is trained on some spreadsheets and tested on new spreadsheets, rather than being transferred to a different task type such as code editing or computer use.
Across our benchmarks, the held-out instances correspond to new spreadsheets or biomedical tasks (SpreadsheetBench, RBioBench), new code repositories (SWE-bench), new medical specialties (HealthBench), or new data families (SciVisAgentBench).
Thus, our notion of generalization is within-task-type generalization, not transfer across task types.
Let $a_q(s)=\frac{1}{|D_q|}\sum_{x\in D_q} r(x,s)$ be the success rate on set $q\in\{\mathrm{train},\mathrm{test}\}$.
Measured against No Skill, in percentage points,
\begin{align}
G_{\mathrm{train}}(s)&=100\,[a_{\mathrm{train}}(s)-a_{\mathrm{train}}(s_\varnothing)],\qquad
G_{\mathrm{test}}(s)=100\,[a_{\mathrm{test}}(s)-a_{\mathrm{test}}(s_\varnothing)],\\
\mathrm{Retention}(s)&=G_{\mathrm{test}}(s)-G_{\mathrm{train}}(s).\label{eq:retention}
\end{align}
The training gain is what the method optimizes; the test gain is what a user receives.
For a skill with $G_{\mathrm{train}}>0$, retention of zero or more means the whole gain carried over, negative retention with $G_{\mathrm{test}}>0$ means part of it did, and $G_{\mathrm{test}}\le 0$ means none did.
A skill with $G_{\mathrm{train}}>0$ is \emph{overfit} when its retention is negative, and \emph{harmful} when $G_{\mathrm{test}}<0$.

\subsection{Analysis}

We run six existing methods and No Skill with one model (Qwen3-Coder-480B) on the six benchmarks of Section~\ref{sec:experimental-design} and ask three questions.
RBioBench Clinical and Omics share one training run; the same final skill is tested on each.

\paragraph{How common is skill overfitting?} Of the 36 final skills from existing methods (six methods on six benchmarks), 21 have $G_{\mathrm{train}}>0$.
Of these 21, five keep the whole gain on test, thirteen keep part of it, and three keep none of it; two of the three are harmful.
Retention ranges from $+3.4$ points (EvoSkill on RBioBench Clinical) to $-15.0$ (EvoSkill on SpreadsheetBench).
Overfitting is common but uneven, and the same method can behave differently on different benchmarks (Figure~\ref{fig:train-test-generalization-case}).
Six benchmarks are too few to tell whether the distance between training and test tasks explains it: HealthBench, which tests on new specialties, is the only benchmark where every skill beats No Skill, while on RBioBench Omics no existing method does.

\paragraph{What do overfit skills contain?} Reading the final skills shows a common pattern (Figure~\ref{fig:skill-content}a): a fix for one training task written as a rule for every task.
One skill fixes column names and an output file as if every task had them; one applies a checklist to every task without saying when it applies; one turns ``create a summary file when one is required'' into ``always create one''.
The round-9 edit in Figure~\ref{fig:story}b is this last change as it happened: training rises by 5 points and test falls by 5.
Not every global rule hurts: the round-2 edit in Figure~\ref{fig:story}b adds general checks and helps.
The harmful rules bind a specific column, file, or output.
These are examples, not a count, but they show how the gap can arise: each edit is fitted to the failures in hand and is stated more broadly than its evidence.

\paragraph{Can the skill content tell us which edits carry over?} An LLM judge that reads only the skill orders finished skills mostly the same way as the test results, but predicts the effect of single edits poorly (Section~\ref{sec:judge-diagnostics}); in Figure~\ref{fig:story}b it calls both edits harmful.
The content of a finished skill says a lot about how well it does on test tasks; whether one edit helps has to be found by running it.

\begin{figure*}[t]
    \centering
    \includegraphics[width=\textwidth]{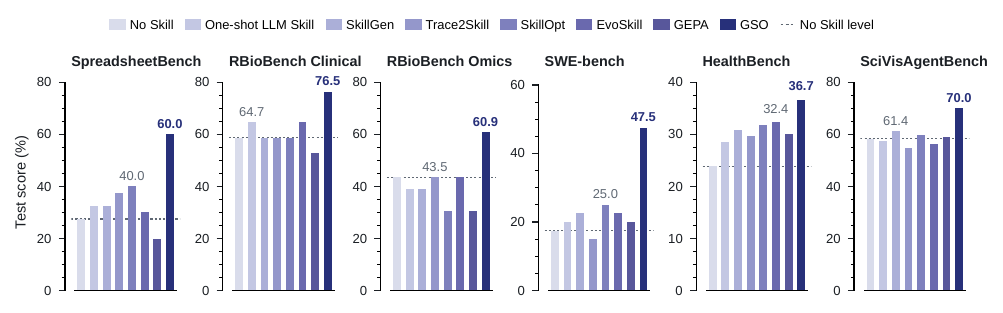}
    \caption{Test score (\%) of final skills on the six benchmarks, with the same model and the same tasks for every method.
        Labels mark \method and the best existing method; the dashed line marks No Skill.
        SciVisAgentBench and HealthBench use family-balanced scores (Section~\ref{sec:experimental-design}).}
    \label{fig:domain-grouped-main-results}
\end{figure*}

\subsection{Problem definition}

In Eq.~(\ref{eq:self-evolve}) the object that is learned and the object that is used on test tasks are the same skill $s$.
Every edit to $s$ is fitted to failures on training tasks, and nothing in the update separates a procedure that carries over from the details of the tasks that caused the failure.
Learning the skill content directly therefore invites overfitting, which is what the analysis finds.
We instead learn how to write a skill.
Let $m$ be a \emph{metaskill}, a guide that tells the model how to write a skill for a given task, and let $C$ be one model call that writes the skill for task $x$: $s_x=C(m,x)$.
The goal is
\begin{equation}
\max_{m}\;\; \mathbb{E}_{x\sim D_{\mathrm{test}}}\big[\,r\big(x,\,C(m,x)\big)\big],
\label{eq:objective}
\end{equation}
estimated during training on $D_{\mathrm{train}}$ and $D_{\mathrm{val}}$.
Only $m$ is kept from task to task; details such as file names, columns, and parameters belong to $s_x$ and are thrown away after task $x$.
Two requirements follow from the analysis.
An edit to $m$ should change as little as possible, so that a fix for one task cannot rewrite the rest; and an edit should be kept only after running it, since its content does not reveal its effect.
In short, we learn how to write skills, not the skill itself.

\section{Generalizable Skill Optimization}
\label{sec:method}

\subsection{Method overview}

Section~\ref{sec:generalization} ends with a change of learning target: instead of the skill $s$ that is used on every task, learn a metaskill $m$ that writes a skill for each task, $s_x=C(m,x)$, and maximize Eq.~(\ref{eq:objective}).
Generalizable Skill Optimization (\method) is one way to do this (Figure~\ref{fig:gso-framework}).
Every step is a call to the same language model that all methods use (Section~\ref{sec:experimental-design}), each with its own prompt.

\paragraph{The metaskill.} The metaskill $m=(m_1,\dots,m_6)$ is a short guide with six parts.
Each part tells the model how to do one step of writing and using a skill: $m_1$ how to read a task, $m_2$ how to choose tools and sources, $m_3$ how to write a step-by-step procedure, $m_4$ how to check the result, $m_5$ how to find what went wrong, and $m_6$ how to repair the output in one retry.
The metaskill never names a specific file, column, or output path; those come from the task.
Every benchmark starts from the same human-written $m^{(0)}$ (Appendix~\ref{app:gso-procedure}).

\paragraph{The skill for one task.} For a task $x$, the model reads $m$ and what the task shows (its prompt, input files, installed tools, and documentation) and writes $s_x=C(m,x)$.
It never sees gold answers, reference solutions, or grader code.
If the task does not say which tool or format to use, $s_x$ says so instead of guessing.
The agent then solves $x$ with $s_x$, checks its own output, and may retry once.
After the task, $s_x$ is thrown away, so a detail that fits $x$ cannot leak into another task.

\paragraph{Two rules for learning $m$.} From Section~\ref{sec:generalization}: an edit changes one part $m_k$ only, and it is kept only if it fixes more than twice as many validation tasks as it breaks.

\begin{table}[t]
\centering
\caption{What the judge scores.
    Each dimension is rated 1--5 from the skill alone; the weighted sum is the judge score.}
\label{tab:judge-rubric}
\small
\begin{tabularx}{\textwidth}{lcX}
\toprule
Dimension & Weight & Question the judge answers \\
\midrule
Generalizability & 30\% & Does the skill describe a reusable procedure, or does it hard-code details of specific tasks? \\
Applicability & 20\% & Does it say when and where its rules apply? \\
Executability & 20\% & Can an agent turn the instructions into concrete actions? \\
Robustness & 20\% & Does it plan for failures without imposing harmful global rules? \\
Clarity/efficiency & 10\% & Is it short enough to follow without losing needed detail? \\
\bottomrule
\end{tabularx}
\end{table}

\subsection{Learning procedure}

Each round $t$ has four steps.

\textit{1. Solve.} For a batch of training tasks, write $s_x=C(m^{(t)},x)$ and run the agent.

\textit{2. Trace the failure.} For a failed task, the model reads the log of the run, finds the first step that went wrong, for example reading the wrong input file or never checking the output, and decides which part $m_k$ should have prevented it.

\textit{3. Edit one part.} The model rewrites $m_k$, giving a candidate $m'$ that differs from $m^{(t)}$ in one part.

\textit{4. Accept or reject.} Run $m^{(t)}$ and $m'$ on the same validation tasks.
A \emph{repair} is a task that $m^{(t)}$ fails and $m'$ passes; a \emph{regression} is a task that $m^{(t)}$ passes and $m'$ fails.
With $R$ and $B$ the sets of repairs and regressions (tasks broken), the update score is
\begin{equation}
u(m',m^{(t)})=|R|-2|B|,
\qquad
m^{(t+1)}=\begin{cases}m' & \text{if } u>0,\\ m^{(t)} & \text{otherwise.}\end{cases}
\label{eq:gate}
\end{equation}
A regression costs twice as much as a repair because a skill that breaks tasks it used to solve is worse for a user than one that stays the same.
The factor 2 was fixed before any reported run, and skill length and cost play no part in the decision.
On SciVisAgentBench and HealthBench, where each task gets a graded score, $u$ is the sum of score increases minus twice the sum of score decreases; HealthBench adds stricter per-specialty checks (Appendix~\ref{app:gso-procedure}).

At test time, nothing is updated: for each test task, \method writes one skill from the final metaskill and runs it.
Edits made on one benchmark are not carried to another.
Compared with Eq.~(\ref{eq:self-evolve}), the two differences are what is learned ($m$ instead of $s$) and what is used on a test task (a fresh $s_x$ instead of one shared $\hat s$).

\section{Experiments and Results}
\label{sec:experimental-design}
\label{sec:results}

\subsection{Setup}

\begin{wrapfigure}{r}{0.45\textwidth}
\vspace{-1.0em}
\centering
\includegraphics[width=0.44\textwidth]{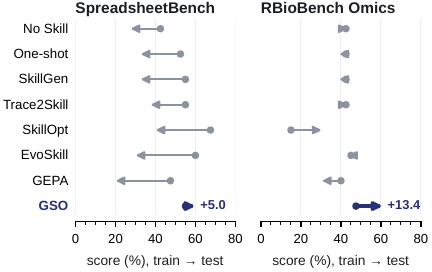}
\caption{Training and test score of each final skill; each arrow goes from training to test (Appendix Table~\ref{tab:main-results}).
    No Skill's own arrow shows how much harder or easier the test tasks are.}
\label{fig:train-test-comparison}
\vspace{-1.0em}
\end{wrapfigure}

We compare eight methods: No Skill; One-shot LLM Skill, a skill the model writes once and never changes; five self-evolving methods (SkillGen~\citep{ma2026skillgen}, Trace2Skill~\citep{ni2026trace2skill}, SkillOpt~\citep{yang2026skillopt}, EvoSkill~\citep{alzubi2026evoskill}, and GEPA~\citep{agrawal2025gepa}); and \method.
All model calls use the same model, Qwen3-Coder-480B-A35B-Instruct-FP8~\citep{yang2025qwen3}, at temperature zero.
Within each benchmark, every method gets the same agent, tasks, feedback, at most 20 rounds, and limits on steps and time per task, and keeps its own rule for when to stop.
Test tasks stay locked until every method has chosen its final skill.
At test time, \method also makes one model call per task to write the skill, checks its output, and may retry once; the other methods do not, so the comparison with \method is between whole methods.

\subsection{Benchmarks}

\begin{figure}[t]
\centering
\includegraphics[width=\textwidth]{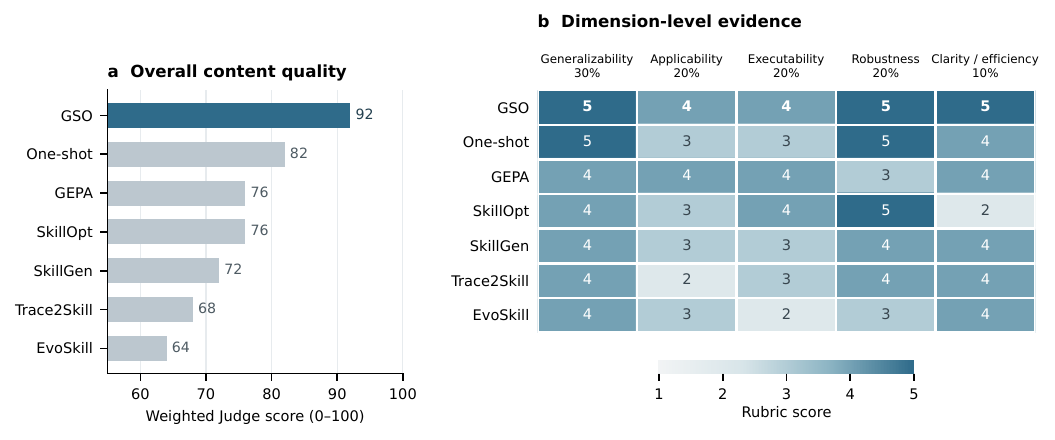}
\caption{Judge scores for the final SpreadsheetBench skills, from the skill alone (questions in Table~\ref{tab:judge-rubric}).
    (a) Weighted score: the \method metaskill scores 92 and the other skills 64 to 82, a loose comparison because the questions were written for skills, not for a guide.
    (b) The five 1--5 dimensions.
    SkillOpt loses on clarity, Trace2Skill on applicability, EvoSkill on executability.}
\label{fig:reasonable-judge}
\end{figure}

We use six benchmarks.
SpreadsheetBench~\citep{ma2024spreadsheetbench} asks the agent to edit spreadsheets.
RBioBench Clinical and RBioBench Omics come from RBioBench, a benchmark we built (Appendix~\ref{app:benchmarks}).
Each task asks the agent to write and run an R program with a real R package, and checks the files it produces against reference outputs.
Clinical tasks use clinical-trial packages such as admiral; Omics tasks use bioinformatics packages such as maftools and Biostrings.
We report the two parts as separate benchmarks; they share one set of training and validation tasks, and only the test tasks differ.
SWE-bench Verified~\citep{jimenez2024swebench,chowdhury2024swebenchverified} (SWE-bench below) asks the agent to fix bugs in real code repositories.
SciVisAgentBench~\citep{ai2026scivisagentbench} asks it to make scientific visualizations.
HealthBench asks it to answer clinicians' health questions from HealthBench Professional~\citep{arora2025healthbench,hicks2026healthbenchpro} using a literature search; we set up this version, and answers are scored with rubrics written by physicians.
RBioBench Clinical and Omics have 17 and 23 test tasks, SciVisAgentBench 20, and the others 40, so one task moves a score by 2.5 to 5.9 points; small differences on one benchmark should be read with care.
Appendix Table~\ref{tab:appendix-benchmark-inventory} gives task counts, splits, and how answers are checked.

\subsection{Scores and the judge}

For most benchmarks, the score is the percentage of tasks the agent solves, as checked by the benchmark's own grader.
SciVisAgentBench and HealthBench give each task a graded score; for these we average over groups of similar tasks so each group counts equally (a family-balanced score).

The judge is an LLM (gpt-5.6-sol) that scores a skill on the five questions in Table~\ref{tab:judge-rubric}.
It sees only the skill, is used only after all runs finish, and never affects which skill a method keeps.

\begin{figure*}[t]
    \centering
    \includegraphics[width=\textwidth]{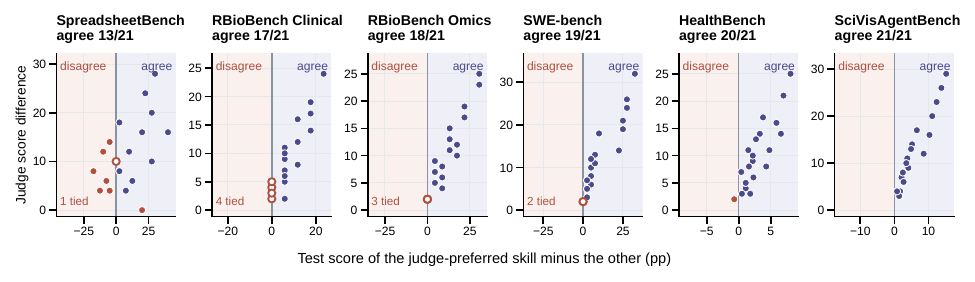}
    \caption{Does the skill the judge prefers also score higher on test?
        Each point is one of the 126 pairs of final skills from the same benchmark, with A the skill the judge prefers.
        The horizontal axis is the test score of A minus B; the vertical axis is the judge score of A minus B. Points to the right agree, points to the left disagree, and hollow points on the zero line are ties on test; ties on test or on judge score count as disagreements.
        No Skill is left out because it has no skill to score.}
    \label{fig:checkpoint-judge-generalization}
\end{figure*}

\subsection{Main results}

Figure~\ref{fig:domain-grouped-main-results} and Appendix Table~\ref{tab:main-results} give the test scores.
Three things stand out.
First, no existing method is best everywhere.
SkillOpt is the best existing method on SpreadsheetBench and SWE-bench, One-shot LLM Skill and EvoSkill (tied) on RBioBench Clinical, SkillGen on SciVisAgentBench, and EvoSkill on HealthBench; on RBioBench Omics none of them beats No Skill.
Second, rewriting a skill can make things worse.
On five of the six benchmarks at least one existing method scores below No Skill, and overall 10 of the 36 test scores of existing methods do.
Third, HealthBench is the only benchmark where every skill helps: on 40 test questions from medical specialties not seen in training, the score rises from 23.9\% with no skill to 28.6--32.4\% for the existing methods and 36.7\% for \method.
\method has the highest test score on all six benchmarks, from 4.3 points (HealthBench) to 22.5 points (SWE-bench) above the best existing method.

Figure~\ref{fig:train-test-comparison} compares test with training scores for the final skills.
On SpreadsheetBench every existing method loses 17.5 to 30 points from training to test; No Skill itself loses 15, so part of each gap comes from harder test tasks, which retention (Eq.~\ref{eq:retention}) removes.

\subsection{What the judge can and cannot see}
\label{sec:judge-diagnostics}

Figure~\ref{fig:reasonable-judge} shows the judge's scores for the final SpreadsheetBench skills.
The skills lose points in different places: SkillOpt for being long and repetitive, Trace2Skill for not saying when its rules apply, and EvoSkill for instructions that are hard to act on.

For finished skills, we take every pair of final skills on the same benchmark (126 pairs) and ask whether the skill with the higher judge score also has the higher test score (Figure~\ref{fig:checkpoint-judge-generalization}).
They agree on 108 pairs (85.7\%); of the other 18, 8 are real disagreements and 10 are ties on test.
These pairs include \method, which is scored on its metaskill, and Clinical and Omics share the same skills, so the pairs are not all independent; agreement is lowest on SpreadsheetBench (13 of 21).
The skills were final when scored, so this is agreement after the fact, not a prediction.

For single edits, we use 45 edits that the existing methods proposed during training (21 on SpreadsheetBench, 24 from the shared RBioBench run).
The judge estimates the chance that an edit fixes a validation task, $p_J(\mathrm{repair})$, and the chance that it breaks one, $p_J(\mathrm{regression})$; we compare $u_J=p_J(\mathrm{repair})-2p_J(\mathrm{regression})$, the same form as the update score of Section~\ref{sec:method}, with the real test change, because the test change is what matters to a user.
The Spearman correlation is 0.39 on SpreadsheetBench and about zero on RBioBench: the content of a finished skill tells a lot about its test score, the content of a single edit little.

\section{Discussion and Conclusion}
\label{sec:discussion}
\label{sec:conclusion}

We find that self-evolving skills keep less of their gain on test tasks than their training scores suggest, and that the amount varies widely: retention ranges from $+3.4$ to $-15.0$ points, and 10 of the 36 skills from existing methods end below No Skill.
Reading the skills, we observe that the ones that carry over worst turn fixes for single training tasks into rules for every task (Figure~\ref{fig:skill-content}), and that whether a single edit helps is hard to read from its content.
We observe that the problem lies in what is learned: when a method learns the skill content itself, each edit fits the training tasks in hand.
Learning how to write skills instead, as \method does, gives the best test score on all six benchmarks; \method itself keeps less than its whole gain on HealthBench (retention $-5.2$), and since it also writes a fresh skill and may retry at test time, this shows that the approach works as a whole, not which part of it causes the gain.

For those who build self-evolving methods, our results suggest three practices.
First, report the test gain and the retention, not only the training curve, because the training score overstates what carries over.
Second, keep an edit only after running it, because a judge that reads the edit predicts its test effect poorly.
Third, use a judge that reads finished skills as a cheap way to rank them; it agrees with the test results on 108 of 126 pairs.
We ran every method with the same model and agent setup by design: skills are now designed and tested under varied settings, and a shared setting keeps the comparison controlled.
Within this setting we studied tasks of one kind.
Open questions are a systematic count of these patterns, whether the findings hold for other models, and whether a metaskill learned on one kind of task helps on another.

\subsection*{Reproducibility Statement}
Section~\ref{sec:experimental-design} reports the model, task partitions,
budgets, verifiers, metrics, and test-access boundary. Appendix~\ref{app:details}
gives the judge rubric, checkpoint policy, result accounting, and figure generation.

\subsection*{AI Use Statement}
In this work, we used generative AI tools for code implementation, feedback on the research methodology (discussing the experimental design), hypothesis refinement, the mathematical formulation of the definitions, data analysis (organizing results), and generating and cleaning candidate RBioBench tasks.
We did not use generative AI tools for translation or for interpreting results, and theoretical model development and proofs are not applicable to this work.
Additionally, we used generative AI tools for literature discovery, figure creation, and drafting and editing the paper.
We have reviewed all AI-assisted work: the authors checked the cited sources, executed and reviewed every experiment, reviewed the RBioBench tasks, whose outputs are checked by executable verifiers against reference outputs, and decided the final methodology and claims.
We take responsibility for the final content of this work, including text, claims, and artifacts produced with generative AI.

\subsection*{Ethics Statement}
This study evaluates language agents on existing benchmark tasks. It does not
deploy systems to users or provide medical advice. HealthBench examples are
used only as released benchmark data within our explicitly identified
retrieval-workflow evaluation.

\bibliography{main}

\begin{thebibliography}{62}
\providecommand{\natexlab}[1]{#1}
\providecommand{\url}[1]{\texttt{#1}}
\expandafter\ifx\csname urlstyle\endcsname\relax
  \providecommand{\doi}[1]{doi: #1}\else
  \providecommand{\doi}{doi: \begingroup \urlstyle{rm}\Url}\fi

\bibitem[Agrawal et~al.(2026)Agrawal, Tan, Soylu, Ziems, Khare, Opsahl-Ong, Singhvi, Shandilya, Ryan, Jiang, Potts, Sen, Dimakis, Stoica, Klein, Zaharia, and Khattab]{agrawal2025gepa}
Lakshya~A. Agrawal, Shangyin Tan, Dilara Soylu, Noah Ziems, Rishi Khare, Krista Opsahl-Ong, Arnav Singhvi, Herumb Shandilya, Michael~J. Ryan, Meng Jiang, Christopher Potts, Koushik Sen, Alexandros~G. Dimakis, Ion Stoica, Dan Klein, Matei Zaharia, and Omar Khattab.
\newblock {GEPA}: Reflective prompt evolution can outperform reinforcement learning.
\newblock In \emph{International Conference on Learning Representations}, 2026.

\bibitem[Ai et~al.(2026)Ai, Miao, Tang, Gorski, Sun, Liu, Ingolfsson, Lenz, Guo, Yu, Leburu, Molash, Wang, Peterka, Wang, and Liu]{ai2026scivisagentbench}
Kuangshi Ai, Haichao Miao, Kaiyuan Tang, Nathaniel Gorski, Jianxin Sun, Guoxi Liu, Helgi~I. Ingolfsson, David Lenz, Hanqi Guo, Hongfeng Yu, Teja Leburu, Michael Molash, Bei Wang, Tom Peterka, Chaoli Wang, and Shusen Liu.
\newblock {SciVisAgentBench}: A benchmark for evaluating scientific data analysis and visualization agents, 2026.
\newblock Accepted to IEEE VIS 2026.

\bibitem[Alzubi et~al.(2026)Alzubi, Provenzano, Bingham, Chen, and Vu]{alzubi2026evoskill}
Salaheddin Alzubi, Noah Provenzano, Jaydon Bingham, Weiyuan Chen, and Tu~Vu.
\newblock {EvoSkill}: Automated skill discovery for multi-agent systems, 2026.

\bibitem[Arora et~al.(2025)Arora, Wei, Hicks, Bowman, Qui{\~n}onero-Candela, Tsimpourlas, Sharman, Shah, Vallone, Beutel, Heidecke, and Singhal]{arora2025healthbench}
Rahul~K. Arora, Jason Wei, Rebecca~Soskin Hicks, Preston Bowman, Joaquin Qui{\~n}onero-Candela, Foivos Tsimpourlas, Michael Sharman, Meghan Shah, Andrea Vallone, Alex Beutel, Johannes Heidecke, and Karan Singhal.
\newblock {HealthBench}: Evaluating large language models towards improved human health, 2025.

\bibitem[Cai et~al.(2024)Cai, Wang, Ma, Chen, and Zhou]{cai2024latm}
Tianle Cai, Xuezhi Wang, Tengyu Ma, Xinyun Chen, and Denny Zhou.
\newblock Large language models as tool makers.
\newblock In \emph{The Twelfth International Conference on Learning Representations}, 2024.

\bibitem[Chen et~al.(2026)Chen, Ye, Yang, Shen, Shen, and Lin]{chen2026cure}
Guirong Chen, Shuqi Ye, Wenkai Yang, Shiqi Shen, Guangyao Shen, and Yankai Lin.
\newblock {CURE}: Critique-driven unified reinforcement learning for test-time self-improvement.
\newblock In \emph{Proceedings of the 64th Annual Meeting of the {A}ssociation for {C}omputational {L}inguistics (Volume 1: Long Papers)}, pp.\  28632--28653, July 2026.
\newblock ISBN 979-8-89176-390-6.

\bibitem[Chen et~al.(2024)Chen, Arkin, Hao, Zhang, Roy, and Fan]{chen2024promst}
Yongchao Chen, Jacob Arkin, Yilun Hao, Yang Zhang, Nicholas Roy, and Chuchu Fan.
\newblock {PR}ompt optimization in multi-step tasks ({PROMST}): Integrating human feedback and heuristic-based sampling.
\newblock In \emph{Proceedings of the 2024 Conference on Empirical Methods in Natural Language Processing}, pp.\  3859--3920, November 2024.

\bibitem[Cheng et~al.(2026)Cheng, Liu, Shan, Wang, Zhu, Ma, Wang, Guo, Lin, and Wang]{cheng2026mem2evolve}
Zihao Cheng, Zeming Liu, Yingyu Shan, Xinyi Wang, Xiangrong Zhu, Yunpu Ma, Hongru Wang, Yuhang Guo, Wei Lin, and Yunhong Wang.
\newblock {Mem$^2$Evolve}: Towards self-evolving agents via co-evolutionary capability expansion and experience distillation.
\newblock In \emph{Proceedings of the 64th Annual Meeting of the Association for Computational Linguistics (Volume 1: Long Papers)}, pp.\  20784--20831, 2026.
\newblock \doi{10.18653/v1/2026.acl-long.952}.

\bibitem[Chowdhury et~al.(2024)Chowdhury, Aung, Shern, Jaffe, Sherburn, Starace, Mays, Dias, Aljubeh, Glaese, Jimenez, Yang, Ho, Patwardhan, Liu, and Madry]{chowdhury2024swebenchverified}
Neil Chowdhury, James Aung, Chan~Jun Shern, Oliver Jaffe, Dane Sherburn, Giulio Starace, Evan Mays, Rachel Dias, Marwan Aljubeh, Mia Glaese, Carlos~E. Jimenez, John Yang, Leyton Ho, Tejal Patwardhan, Kevin Liu, and Aleksander Madry.
\newblock Introducing {SWE}-bench {Verified}.
\newblock OpenAI blog, \url{https://openai.com/index/introducing-swe-bench-verified/}, August 2024.

\bibitem[Fang et~al.(2026)Fang, Liang, Wang, Wu, Qiao, Xie, Huang, Chen, and Zhang]{fang2026memp}
Runnan Fang, Yuan Liang, Xiaobin Wang, Jialong Wu, Shuofei Qiao, Pengjun Xie, Fei Huang, Huajun Chen, and Ningyu Zhang.
\newblock Memp: Exploring agent procedural memory.
\newblock In \emph{Findings of the Association for Computational Linguistics: ACL 2026}, pp.\  17490--17502, 2026.
\newblock \doi{10.18653/v1/2026.findings-acl.866}.

\bibitem[Fu et~al.(2026)Fu, Mei, Wu, Yang, Xu, Wang, Cai, Liu, Wen, and Shi]{fu2026agentfirstday}
Daocheng Fu, Jianbiao Mei, Rong Wu, Xuemeng Yang, Jia Xu, Ding Wang, Pinlong Cai, Yong Liu, Licheng Wen, and Botian Shi.
\newblock The agent{'}s first day: Benchmarking learning, exploration, and scheduling in the workplace scenarios.
\newblock In \emph{Findings of the {A}ssociation for {C}omputational {L}inguistics: {ACL} 2026}, pp.\  30094--30109, July 2026.
\newblock ISBN 979-8-89176-395-1.

\bibitem[Guo et~al.(2026)Guo, Zhou, Liu, Qian, Ren, Shao, Fan, Fung, Wang, Zhang, and Shao]{guo2026trace}
Dadi Guo, Tianyi Zhou, Dongrui Liu, Chen Qian, Qihan Ren, Shuai Shao, Zhiyuan Fan, Yi~R. Fung, Kun Wang, Linfeng Zhang, and Jing Shao.
\newblock Towards self-evolving agent benchmarks: Validatable agent trajectory via test-time exploration.
\newblock In \emph{The Fourteenth International Conference on Learning Representations}, 2026.

\bibitem[Guo et~al.(2024)Guo, Wang, Guo, Li, Song, Tan, Liu, Bian, and Yang]{guo2024evoprompt}
Qingyan Guo, Rui Wang, Junliang Guo, Bei Li, Kaitao Song, Xu~Tan, Guoqing Liu, Jiang Bian, and Yujiu Yang.
\newblock Connecting large language models with evolutionary algorithms yields powerful prompt optimizers.
\newblock In \emph{The Twelfth International Conference on Learning Representations}, 2024.

\bibitem[Gupta et~al.(2024)Gupta, Kirtania, Singha, Gulwani, Radhakrishna, Soares, and Shi]{gupta2024metareflection}
Priyanshu Gupta, Shashank Kirtania, Ananya Singha, Sumit Gulwani, Arjun Radhakrishna, Gustavo Soares, and Sherry Shi.
\newblock {M}eta{R}eflection: Learning instructions for language agents using past reflections.
\newblock In \emph{Proceedings of the 2024 Conference on Empirical Methods in Natural Language Processing}, pp.\  8369--8385, November 2024.

\bibitem[Hao et~al.(2026)Hao, Wang, Luo, Zhang, Zhou, Lin, Wang, Dong, and Chen]{hao2026recreate}
Zhezheng Hao, Hong Wang, Jian Luo, Jianqing Zhang, Yuyan Zhou, Qiang Lin, Can Wang, Hande Dong, and Jiawei Chen.
\newblock {R}e{C}reate: Reasoning and creating domain agents driven by experience.
\newblock In \emph{Proceedings of the 64th Annual Meeting of the {A}ssociation for {C}omputational {L}inguistics (Volume 1: Long Papers)}, pp.\  31018--31046, July 2026.
\newblock ISBN 979-8-89176-390-6.

\bibitem[He et~al.(2025)He, Li, Chen, Liu, Chen, Sui, Chen, Zhu, Luo, Yang, and Hooi]{he2025hitl}
Yufei He, Ruoyu Li, Alex Chen, Yue Liu, Yulin Chen, Yuan Sui, Cheng Chen, Yi~Zhu, Luca Luo, Frank Yang, and Bryan Hooi.
\newblock Enabling self-improving agents to learn at test time with human-in-the-loop guidance.
\newblock In \emph{Proceedings of the 2025 Conference on Empirical Methods in Natural Language Processing: Industry Track}, pp.\  1625--1653, November 2025.
\newblock ISBN 979-8-89176-333-3.
\newblock \doi{10.18653/v1/2025.emnlp-industry.115}.

\bibitem[Hicks et~al.(2026)Hicks, Trofimov, Lim, Arora, Tsimpourlas, Bowman, Sharman, Tong, Karthik, Dugar, Jagadeesh, Saab, Heidecke, Alexander, Gross, and Singhal]{hicks2026healthbenchpro}
Rebecca~Soskin Hicks, Mikhail Trofimov, Dominick Lim, Rahul~K. Arora, Foivos Tsimpourlas, Preston Bowman, Michael Sharman, Chi Tong, Kavin Karthik, Arnav Dugar, Akshay Jagadeesh, Khaled Saab, Johannes Heidecke, Ashley Alexander, Nate Gross, and Karan Singhal.
\newblock {HealthBench} professional: Evaluating large language models on real clinician chats, 2026.

\bibitem[Hu et~al.(2025)Hu, Lu, and Clune]{hu2024adas}
Shengran Hu, Cong Lu, and Jeff Clune.
\newblock Automated design of agentic systems.
\newblock In \emph{International Conference on Learning Representations}, 2025.

\bibitem[Jiang et~al.(2025)Jiang, Lin, Cao, Tian, Kang, Wang, Sun, and Han]{jiang2025deepretrieval}
Pengcheng Jiang, Jiacheng Lin, Lang Cao, Runchu Tian, SeongKu Kang, Zifeng Wang, Jimeng Sun, and Jiawei Han.
\newblock {DeepRetrieval}: Hacking real search engines and retrievers with large language models via reinforcement learning, 2025.

\bibitem[Jimenez et~al.(2024)Jimenez, Yang, Wettig, Yao, Pei, Press, and Narasimhan]{jimenez2024swebench}
Carlos~E. Jimenez, John Yang, Alexander Wettig, Shunyu Yao, Kexin Pei, Ofir Press, and Karthik Narasimhan.
\newblock {SWE}-bench: Can language models resolve real-world {GitHub} issues?
\newblock In \emph{The Twelfth International Conference on Learning Representations}, 2024.

\bibitem[Jin et~al.(2025)Jin, Zeng, Yue, Yoon, Arik, Wang, Zamani, and Han]{jin2025searchr1}
Bowen Jin, Hansi Zeng, Zhenrui Yue, Jinsung Yoon, Sercan Arik, Dong Wang, Hamed Zamani, and Jiawei Han.
\newblock {Search-R1}: Training {LLM}s to reason and leverage search engines with reinforcement learning, 2025.

\bibitem[Khattab et~al.(2024)Khattab, Singhvi, Maheshwari, Zhang, Santhanam, A, Haq, Sharma, Joshi, Moazam, Miller, Zaharia, and Potts]{khattab2024dspy}
Omar Khattab, Arnav Singhvi, Paridhi Maheshwari, Zhiyuan Zhang, Keshav Santhanam, Sri~Vardhamanan A, Saiful Haq, Ashutosh Sharma, Thomas Joshi, Hanna Moazam, Heather Miller, Matei Zaharia, and Christopher Potts.
\newblock {DSPy}: Compiling declarative language model calls into state-of-the-art pipelines.
\newblock In \emph{The Twelfth International Conference on Learning Representations}, 2024.

\bibitem[Liang et~al.(2026)Liang, Cao, Zhao, Teng, Liao, Zhao, and Liu]{liang2026remember}
Sirui Liang, Pengfei Cao, Jian Zhao, Wenhao Teng, Xiangwen Liao, Jun Zhao, and Kang Liu.
\newblock Learning how to remember: A meta-cognitive management method for structured and transferable agent memory.
\newblock In \emph{Findings of the {A}ssociation for {C}omputational {L}inguistics: {ACL} 2026}, pp.\  30733--30753, July 2026.
\newblock ISBN 979-8-89176-395-1.

\bibitem[Ling et~al.(2026)Ling, Liao, Jiang, and Guan]{ling2026reusable}
Shuai Ling, Lizi Liao, Dongmei Jiang, and Weili Guan.
\newblock Reusable experiences: Latent routing and modular composition in {LLM}s.
\newblock In \emph{Proceedings of the 64th Annual Meeting of the {A}ssociation for {C}omputational {L}inguistics (Volume 1: Long Papers)}, pp.\  30087--30100, July 2026.
\newblock ISBN 979-8-89176-390-6.

\bibitem[Liu et~al.(2025)Liu, Si, Narasimhan, and Yao]{liu2025cer}
Yitao Liu, Chenglei Si, Karthik~R Narasimhan, and Shunyu Yao.
\newblock Contextual experience replay for self-improvement of language agents.
\newblock In \emph{Proceedings of the 63rd Annual Meeting of the Association for Computational Linguistics (Volume 1: Long Papers)}, pp.\  14179--14198, July 2025.
\newblock ISBN 979-8-89176-251-0.

\bibitem[Ma et~al.(2026{\natexlab{a}})Ma, Ma, Lin, Yan, Bi, Cao, Tian, Tresp, and Schuetze]{ma2026textualbackprop}
Xiaowen Ma, Yunpu Ma, Chenyang Lin, Sikuan Yan, Jinhe Bi, Zixuan Cao, Yijun Tian, Volker Tresp, and Hinrich Schuetze.
\newblock Self-evolving multi-agent systems via textual backpropagation.
\newblock In \emph{Findings of the {A}ssociation for {C}omputational {L}inguistics: {ACL} 2026}, pp.\  9918--9951, July 2026{\natexlab{a}}.
\newblock ISBN 979-8-89176-395-1.

\bibitem[Ma et~al.(2026{\natexlab{b}})Ma, Huang, Bao, Zhuang, Shukla, Galley, Zhang, and Feuerriegel]{ma2026skillgen}
Yuchen Ma, Yue Huang, Han Bao, Haomin Zhuang, Swadheen Shukla, Michel Galley, Xiangliang Zhang, and Stefan Feuerriegel.
\newblock {SkillGen}: Verified inference-time agent skill synthesis, 2026{\natexlab{b}}.

\bibitem[Ma et~al.(2024)Ma, Zhang, Zhang, Yu, Zhang, Zhang, Luo, Wang, and Tang]{ma2024spreadsheetbench}
Zeyao Ma, Bohan Zhang, Jing Zhang, Jifan Yu, Xiaokang Zhang, Xiaohan Zhang, Sijia Luo, Xi~Wang, and Jie Tang.
\newblock {SpreadsheetBench}: Towards challenging real world spreadsheet manipulation.
\newblock In \emph{Advances in Neural Information Processing Systems Datasets and Benchmarks Track}, 2024.

\bibitem[Ni et~al.(2026)Ni, Liu, Liu, Sun, Zhou, Cheng, Wang, Zhao, Jiang, and Jiang]{ni2026trace2skill}
Jingwei Ni, Yihao Liu, Xinpeng Liu, Yutao Sun, Mengyu Zhou, Pengyu Cheng, Dexin Wang, Erchao Zhao, Xiaoxi Jiang, and Guanjun Jiang.
\newblock {Trace2Skill}: Distill trajectory-local lessons into transferable agent skills, 2026.

\bibitem[Opsahl-Ong et~al.(2024)Opsahl-Ong, Ryan, Purtell, Broman, Potts, Zaharia, and Khattab]{opsahlong2024mipro}
Krista Opsahl-Ong, Michael~J Ryan, Josh Purtell, David Broman, Christopher Potts, Matei Zaharia, and Omar Khattab.
\newblock Optimizing instructions and demonstrations for multi-stage language model programs.
\newblock In \emph{Proceedings of the 2024 Conference on Empirical Methods in Natural Language Processing}, pp.\  9340--9366, November 2024.

\bibitem[Ouyang et~al.(2026)Ouyang, Yan, Hsu, Chen, Jiang, Wang, Han, Le, Daruki, Tang, Tirumalashetty, Lee, Rofouei, Lin, Han, Lee, and Pfister]{ouyang2026reasoningbank}
Siru Ouyang, Jun Yan, I-Hung Hsu, Yanfei Chen, Ke~Jiang, Zifeng Wang, Rujun Han, Long Le, Samira Daruki, Xiangru Tang, Vishy Tirumalashetty, George Lee, Mahsan Rofouei, Hangfei Lin, Jiawei Han, Chen-Yu Lee, and Tomas Pfister.
\newblock {ReasoningBank}: Scaling agent self-evolving with reasoning memory.
\newblock In \emph{The Fourteenth International Conference on Learning Representations}, 2026.

\bibitem[Park et~al.(2023)Park, O'Brien, Cai, Morris, Liang, and Bernstein]{park2023generative}
Joon~Sung Park, Joseph~C. O'Brien, Carrie~J. Cai, Meredith~Ringel Morris, Percy Liang, and Michael~S. Bernstein.
\newblock Generative agents: Interactive simulacra of human behavior.
\newblock In \emph{Proceedings of the 36th Annual ACM Symposium on User Interface Software and Technology}, 2023.
\newblock \doi{10.1145/3586183.3606763}.

\bibitem[Pryzant et~al.(2023)Pryzant, Iter, Li, Lee, Zhu, and Zeng]{pryzant2023protegi}
Reid Pryzant, Dan Iter, Jerry Li, Yin~Tat Lee, Chenguang Zhu, and Michael Zeng.
\newblock Automatic prompt optimization with ``gradient descent'' and beam search.
\newblock In \emph{Proceedings of the 2023 Conference on Empirical Methods in Natural Language Processing}, pp.\  7957--7968, 2023.
\newblock \doi{10.18653/v1/2023.emnlp-main.494}.

\bibitem[Qian et~al.(2023)Qian, Han, Fung, Qin, Liu, and Ji]{qian2023creator}
Cheng Qian, Chi Han, Yi~Fung, Yujia Qin, Zhiyuan Liu, and Heng Ji.
\newblock {CREATOR}: Tool creation for disentangling abstract and concrete reasoning of large language models.
\newblock In \emph{Findings of the Association for Computational Linguistics: EMNLP 2023}, pp.\  6922--6939, December 2023.

\bibitem[Qu et~al.(2024)Qu, Zhang, Garg, and Kumar]{qu2024rise}
Yuxiao Qu, Tianjun Zhang, Naman Garg, and Aviral Kumar.
\newblock Recursive introspection: Teaching language model agents how to self-improve.
\newblock In \emph{Advances in Neural Information Processing Systems}, volume~37, 2024.

\bibitem[Shang et~al.(2025)Shang, Li, Zhao, Ma, Liu, Xu, and Li]{shang2025agentsquare}
Yu~Shang, Yu~Li, Keyu Zhao, Likai Ma, Jiahe Liu, Fengli Xu, and Yong Li.
\newblock {AgentSquare}: Automatic {LLM} agent search in modular design space.
\newblock In \emph{The Thirteenth International Conference on Learning Representations}, 2025.

\bibitem[Sharma et~al.(2022)Sharma, Torralba, and Andreas]{sharma2022skill}
Pratyusha Sharma, Antonio Torralba, and Jacob Andreas.
\newblock Skill induction and planning with latent language.
\newblock In \emph{Proceedings of the 60th Annual Meeting of the Association for Computational Linguistics (Volume 1: Long Papers)}, pp.\  1713--1726, May 2022.

\bibitem[Shi et~al.(2025)Shi, Ma, Liang, Diao, Ma, and Vosoughi]{shi2024judging}
Lin Shi, Chiyu Ma, Wenhua Liang, Xingjian Diao, Weicheng Ma, and Soroush Vosoughi.
\newblock Judging the judges: A systematic study of position bias in {LLM}-as-a-judge.
\newblock In \emph{Proceedings of the 14th International Joint Conference on Natural Language Processing and the 4th Conference of the Asia-Pacific Chapter of the Association for Computational Linguistics}, pp.\  292--314, 2025.
\newblock \doi{10.18653/v1/2025.ijcnlp-long.18}.

\bibitem[Shinn et~al.(2023)Shinn, Cassano, Gopinath, Narasimhan, and Yao]{shinn2023reflexion}
Noah Shinn, Federico Cassano, Ashwin Gopinath, Karthik Narasimhan, and Shunyu Yao.
\newblock Reflexion: Language agents with verbal reinforcement learning.
\newblock In \emph{Advances in Neural Information Processing Systems}, volume~36, pp.\  8634--8652, 2023.

\bibitem[Suzgun et~al.(2026)Suzgun, Yuksekgonul, Bianchi, Jurafsky, and Zou]{suzgun2026dynamic}
Mirac Suzgun, Mert Yuksekgonul, Federico Bianchi, Dan Jurafsky, and James Zou.
\newblock Dynamic cheatsheet: Test-time learning with adaptive memory.
\newblock In \emph{Proceedings of the 19th Conference of the European Chapter of the Association for Computational Linguistics (Volume 1: Long Papers)}, pp.\  7080--7106, 2026.
\newblock \doi{10.18653/v1/2026.eacl-long.333}.

\bibitem[Wang et~al.(2024)Wang, Xie, Jiang, Mandlekar, Xiao, Zhu, Fan, and Anandkumar]{wang2023voyager}
Guanzhi Wang, Yuqi Xie, Yunfan Jiang, Ajay Mandlekar, Chaowei Xiao, Yuke Zhu, Linxi Fan, and Anima Anandkumar.
\newblock Voyager: An open-ended embodied agent with large language models.
\newblock \emph{Transactions on Machine Learning Research}, 2024.

\bibitem[Wang et~al.(2025{\natexlab{a}})Wang, Guo, Ma, and Zhang]{wang2025experience}
Jiayin Wang, Zhiqiang Guo, Weizhi Ma, and Min Zhang.
\newblock How far can {LLM}s improve from experience? measuring test-time learning ability in {LLM}s with human comparison.
\newblock In \emph{Proceedings of the 2025 Conference on Empirical Methods in Natural Language Processing}, pp.\  25677--25691, 2025{\natexlab{a}}.
\newblock \doi{10.18653/v1/2025.emnlp-main.1304}.

\bibitem[Wang et~al.(2026{\natexlab{a}})Wang, Zhou, Fu, Wang, Liu, Zhang, and Lin]{wang2026skilltta}
Jingxing Wang, Chenyu Zhou, Zhihui Fu, Jun Wang, Weiwen Liu, Weinan Zhang, and Jianghao Lin.
\newblock Skills on the fly: Test-time adaptive skill synthesis for {LLM} agents, 2026{\natexlab{a}}.

\bibitem[Wang et~al.(2026{\natexlab{b}})Wang, Yan, Wang, Tian, Mishra, Xu, Gandhi, Xu, and Cheong]{wang2026sage}
Jiongxiao Wang, Qiaojing Yan, Yawei Wang, Yijun Tian, Soumya~Smruti Mishra, Zhichao Xu, Megha Gandhi, Panpan Xu, and Lin~Lee Cheong.
\newblock Reinforcement learning for self-improving agent with skill library.
\newblock In \emph{Proceedings of the 64th Annual Meeting of the {A}ssociation for {C}omputational {L}inguistics (Volume 1: Long Papers)}, pp.\  1529--1550, July 2026{\natexlab{b}}.
\newblock ISBN 979-8-89176-390-6.

\bibitem[Wang et~al.(2025{\natexlab{b}})Wang, Mao, Fried, and Neubig]{wang2024awm}
Zora~Zhiruo Wang, Jiayuan Mao, Daniel Fried, and Graham Neubig.
\newblock Agent workflow memory.
\newblock In \emph{Proceedings of the 42nd International Conference on Machine Learning}, volume 267 of \emph{Proceedings of Machine Learning Research}, pp.\  63897--63911. PMLR, 2025{\natexlab{b}}.

\bibitem[W{\"o}lflein et~al.(2025)W{\"o}lflein, Ferber, Truhn, Arandjelovic, and Kather]{wolflein2025toolmaker}
Georg W{\"o}lflein, Dyke Ferber, Daniel Truhn, Ognjen Arandjelovic, and Jakob~Nikolas Kather.
\newblock {LLM} agents making agent tools.
\newblock In \emph{Proceedings of the 63rd Annual Meeting of the Association for Computational Linguistics (Volume 1: Long Papers)}, pp.\  26092--26130, July 2025.
\newblock ISBN 979-8-89176-251-0.

\bibitem[Xin et~al.(2026)Xin, Li, Liu, Yan, Wang, Yang, Gu, Yu, and Sun]{xin2026metamem}
Haidong Xin, Xinze Li, Zhenghao Liu, Yukun Yan, Shuo Wang, Cheng Yang, Yu~Gu, Ge~Yu, and Maosong Sun.
\newblock {M}eta{M}em: Evolving meta-memory for knowledge utilization through self-reflective symbolic optimization.
\newblock In \emph{Findings of the {A}ssociation for {C}omputational {L}inguistics: {ACL} 2026}, pp.\  5473--5492, July 2026.
\newblock ISBN 979-8-89176-395-1.
\newblock \doi{10.18653/v1/2026.findings-acl.270}.

\bibitem[Xiong et~al.(2026)Xiong, Lin, Xie, He, Liu, Tang, Lakkaraju, and Xiang]{xiong2026memorymanagement}
Zidi Xiong, Yuping Lin, Wenya Xie, Pengfei He, Zirui Liu, Jiliang Tang, Himabindu Lakkaraju, and Zhen Xiang.
\newblock How memory management impacts {LLM} agents: An empirical study of experience-following behavior.
\newblock In \emph{Proceedings of the 64th Annual Meeting of the Association for Computational Linguistics (Volume 1: Long Papers)}, pp.\  623--645, 2026.
\newblock \doi{10.18653/v1/2026.acl-long.27}.

\bibitem[Yan et~al.(2026)Yan, Yang, Huang, Nie, Ding, Li, Ma, Bi, Kersting, Pan, Schuetze, Tresp, and Ma]{yan2026memoryr1}
Sikuan Yan, Xiufeng Yang, Zuchao Huang, Ercong Nie, Zifeng Ding, Zonggen Li, Xiaowen Ma, Jinhe Bi, Kristian Kersting, Jeff~Z. Pan, Hinrich Schuetze, Volker Tresp, and Yunpu Ma.
\newblock Memory-r1: Enhancing large language model agents to manage and utilize memories via reinforcement learning.
\newblock In \emph{Proceedings of the 64th Annual Meeting of the {A}ssociation for {C}omputational {L}inguistics (Volume 1: Long Papers)}, pp.\  12805--12825, July 2026.
\newblock ISBN 979-8-89176-390-6.

\bibitem[Yang et~al.(2025)Yang, Li, Yang, Zhang, Hui, Zheng, Yu, Gao, et~al.]{yang2025qwen3}
An~Yang, Anfeng Li, Baosong Yang, Beichen Zhang, Binyuan Hui, Bo~Zheng, Bowen Yu, Chang Gao, et~al.
\newblock Qwen3 technical report, 2025.

\bibitem[Yang et~al.(2026{\natexlab{a}})Yang, Yang, Wen, Fu, Mei, Wu, Cai, Shen, Deng, Xu, Shi, Qiao, and Li]{yang2026muse}
Cheng Yang, Xuemeng Yang, Licheng Wen, Daocheng Fu, Jianbiao Mei, Rong Wu, Pinlong Cai, Yufan Shen, Nianchen Deng, Jia Xu, Botian Shi, Yu~Qiao, and Haifeng Li.
\newblock Towards self-evolving agents: Enabling autonomy through interactive experience refinement.
\newblock In \emph{Findings of the Association for Computational Linguistics: ACL 2026}, pp.\  30424--30451, 2026{\natexlab{a}}.
\newblock \doi{10.18653/v1/2026.findings-acl.1522}.

\bibitem[Yang et~al.(2024{\natexlab{a}})Yang, Wang, Lu, Liu, Le, Zhou, and Chen]{yang2023opro}
Chengrun Yang, Xuezhi Wang, Yifeng Lu, Hanxiao Liu, Quoc~V. Le, Denny Zhou, and Xinyun Chen.
\newblock Large language models as optimizers.
\newblock In \emph{International Conference on Learning Representations}, 2024{\natexlab{a}}.

\bibitem[Yang et~al.(2024{\natexlab{b}})Yang, Jimenez, Wettig, Lieret, Yao, Narasimhan, and Press]{yang2024sweagent}
John Yang, Carlos~E. Jimenez, Alexander Wettig, Kilian Lieret, Shunyu Yao, Karthik Narasimhan, and Ofir Press.
\newblock {SWE}-agent: Agent-computer interfaces enable automated software engineering.
\newblock In \emph{Advances in Neural Information Processing Systems}, volume~37, 2024{\natexlab{b}}.

\bibitem[Yang et~al.(2026{\natexlab{b}})Yang, Gong, Huang, Yang, Zhou, Huang, Li, Gao, Dai, Liu, Qiu, Yang, Chen, Yang, and Luo]{yang2026skillopt}
Yifan Yang, Ziyang Gong, Weiquan Huang, Qihao Yang, Ziwei Zhou, Zisu Huang, Yan Li, Xuemei Gao, Qi~Dai, Bei Liu, Kai Qiu, Yuqing Yang, Dongdong Chen, Xue Yang, and Chong Luo.
\newblock {SkillOpt}: Executive strategy for self-evolving agent skills, 2026{\natexlab{b}}.

\bibitem[Yin et~al.(2025)Yin, Wang, Pan, Lin, Wan, and Wang]{yin2025godel}
Xunjian Yin, Xinyi Wang, Liangming Pan, Li~Lin, Xiaojun Wan, and William~Yang Wang.
\newblock G{\"o}del agent: A self-referential agent framework for recursively self-improvement.
\newblock In \emph{Proceedings of the 63rd Annual Meeting of the Association for Computational Linguistics (Volume 1: Long Papers)}, pp.\  27890--27913, 2025.
\newblock \doi{10.18653/v1/2025.acl-long.1354}.

\bibitem[Yuksekgonul et~al.(2025)Yuksekgonul, Bianchi, Boen, Liu, Lu, Huang, Guestrin, and Zou]{yuksekgonul2025textgrad}
Mert Yuksekgonul, Federico Bianchi, Joseph Boen, Sheng Liu, Pan Lu, Zhi Huang, Carlos Guestrin, and James Zou.
\newblock Optimizing generative {AI} by backpropagating language model feedback.
\newblock \emph{Nature}, 639:\penalty0 609--616, 2025.

\bibitem[Zelikman et~al.(2024)Zelikman, Lorch, Mackey, and Kalai]{zelikman2024stop}
Eric Zelikman, Eliana Lorch, Lester Mackey, and Adam~Tauman Kalai.
\newblock Self-taught optimizer ({STOP}): Recursively self-improving code generation.
\newblock In \emph{Conference on Language Modeling}, 2024.

\bibitem[Zhang et~al.(2026{\natexlab{a}})Zhang, Hu, Lu, Lange, and Clune]{zhang2025dgm}
Jenny Zhang, Shengran Hu, Cong Lu, Robert Lange, and Jeff Clune.
\newblock Darwin g{\"o}del machine: Open-ended evolution of self-improving agents.
\newblock In \emph{International Conference on Learning Representations}, 2026{\natexlab{a}}.

\bibitem[Zhang et~al.(2025)Zhang, Xiang, Yu, Teng, Chen, Chen, Zhuge, Cheng, Hong, Wang, Zheng, Liu, Luo, and Wu]{zhang2025aflow}
Jiayi Zhang, Jinyu Xiang, Zhaoyang Yu, Fengwei Teng, XiongHui Chen, Jiaqi Chen, Mingchen Zhuge, Xin Cheng, Sirui Hong, Jinlin Wang, Bingnan Zheng, Bang Liu, Yuyu Luo, and Chenglin Wu.
\newblock {AFlow}: Automating agentic workflow generation.
\newblock In \emph{The Thirteenth International Conference on Learning Representations}, 2025.

\bibitem[Zhang et~al.(2026{\natexlab{b}})Zhang, Lu, Qian, He, and Liu]{zhang2026agentfactory}
Zhang Zhang, Shuqi Lu, Hongjin Qian, Di~He, and Zheng Liu.
\newblock {A}gent{F}actory: A self-evolving framework through executable subagent accumulation and reuse.
\newblock In \emph{Proceedings of the 64th Annual Meeting of the {A}ssociation for {C}omputational {L}inguistics (Volume 3: System Demonstrations)}, pp.\  819--828, July 2026{\natexlab{b}}.
\newblock ISBN 979-8-89176-392-0.

\bibitem[Zhao et~al.(2024)Zhao, Huang, Xu, Lin, Liu, and Huang]{zhao2024expel}
Andrew Zhao, Daniel Huang, Quentin Xu, Matthieu Lin, Yong-Jin Liu, and Gao Huang.
\newblock {ExpeL}: {LLM} agents are experiential learners.
\newblock In \emph{Proceedings of the AAAI Conference on Artificial Intelligence}, volume~38, pp.\  19632--19642, 2024.
\newblock \doi{10.1609/aaai.v38i17.29936}.

\bibitem[Zheng et~al.(2023)Zheng, Chiang, Sheng, Zhuang, Wu, Zhuang, Lin, Li, Li, Xing, Zhang, Gonzalez, and Stoica]{zheng2023mtbench}
Lianmin Zheng, Wei-Lin Chiang, Ying Sheng, Siyuan Zhuang, Zhanghao Wu, Yonghao Zhuang, Zi~Lin, Zhuohan Li, Dacheng Li, Eric~P. Xing, Hao Zhang, Joseph~E. Gonzalez, and Ion Stoica.
\newblock Judging {LLM}-as-a-judge with {MT-Bench} and chatbot arena.
\newblock In \emph{Advances in Neural Information Processing Systems}, volume~36, pp.\  46595--46623, 2023.

\end{thebibliography}
\bibliographystyle{preprint}

\appendix
\section{Supplementary Material}
\label{app:details}

\begingroup \renewcommand{\contentsname}{Appendix contents}
\small
\etocsetnexttocdepth{subsection} \localtableofcontents \endgroup

This appendix expands the experimental protocol, evaluation definitions, and evidence roles used in the main paper.
It distinguishes information available during optimization from analyses performed after the final artifact was frozen.

\begin{table*}[t]
\centering
\caption{Main results (\%).
    Task benchmarks report accuracy; HealthBench reports the family-balanced physician-rubric score, and SciVisAgentBench reports family-balanced official evaluator score.
    Clinical and Omics are two author-constructed benchmarks in the RBioBench suite and share its task format and verifier.
    The two panels split the six reporting columns for readability; values are unchanged.}
\label{tab:main-results}
\small
\setlength{\tabcolsep}{4.5pt}
\begin{tabular}{lccccccccc}
\toprule
\multicolumn{10}{l}{\textbf{Panel A: Spreadsheet and RBioBench}} \\
& \multicolumn{3}{c}{SpreadsheetBench} &
\multicolumn{3}{c}{RBio Clinical} &
\multicolumn{3}{c}{RBio Omics} \\
\cmidrule(lr){2-4}\cmidrule(lr){5-7}\cmidrule(lr){8-10}
Method & Train & Val. & Test & Train & Val. & Test & Train & Val. & Test \\
\midrule
% BEGIN MAIN RESULTS PANEL A
No Skill & 42.5 & 30.0 & 27.5 & 42.5 & 45.0 & 58.8 & 42.5 & 45.0 & 43.5 \\
One-shot LLM Skill & 52.5 & 50.0 & 32.5 & 42.5 & 45.0 & 64.7 & 42.5 & 45.0 & 39.1 \\
SkillGen & 55.0 & 40.0 & 32.5 & 42.5 & 50.0 & 58.8 & 42.5 & 50.0 & 39.1 \\
Trace2Skill & 55.0 & 35.0 & 37.5 & 42.5 & 40.0 & 58.8 & 42.5 & 40.0 & 43.5 \\
SkillOpt & 67.5 & 35.0 & 40.0 & 15.0 & 25.0 & 58.8 & 15.0 & 25.0 & 30.4 \\
EvoSkill & 60.0 & 30.0 & 30.0 & 45.0 & 45.0 & 64.7 & 45.0 & 45.0 & 43.5 \\
GEPA & 47.5 & 30.0 & 20.0 & 40.0 & 40.0 & 52.9 & 40.0 & 40.0 & 30.4 \\
\textbf{\method} & 55.0 & 55.0 & \textbf{60.0} & 47.5 & 50.0 & \textbf{76.5} & 47.5 & 50.0 & \textbf{60.9} \\
% END MAIN RESULTS PANEL A
\midrule
\multicolumn{10}{l}{\textbf{Panel B: Software, medical retrieval, and scientific visualization}} \\
& \multicolumn{3}{c}{SWE-bench Verified} &
\multicolumn{3}{c}{HealthBench} &
\multicolumn{3}{c}{SciVisAgentBench} \\
\cmidrule(lr){2-4}\cmidrule(lr){5-7}\cmidrule(lr){8-10}
Method & Train & Val. & Test & Train & Val. & Test & Train & Val. & Test \\
\midrule
% BEGIN MAIN RESULTS PANEL B
No Skill & 30.0 & 35.0 & 17.5 & 27.8 & 25.4 & 23.9 & 60.7 & 63.9 & 58.3 \\
One-shot LLM Skill & 37.5 & 35.0 & 20.0 & 32.1 & 30.4 & 28.6 & 59.9 & 67.4 & 57.6 \\
SkillGen & 37.5 & 40.0 & 22.5 & 34.6 & 32.7 & 30.8 & 60.7 & 63.9 & \textbf{61.4} \\
Trace2Skill & 32.5 & 25.0 & 15.0 & 33.4 & 31.5 & 29.7 & 60.5 & 67.3 & 54.8 \\
SkillOpt & 40.0 & 35.0 & 25.0 & 37.8 & 34.5 & 31.9 & 59.4 & 66.2 & 59.7 \\
EvoSkill & 40.0 & 40.0 & 22.5 & 37.1 & 34.8 & 32.4 & 59.9 & 66.7 & 56.2 \\
GEPA & 37.5 & 45.0 & 20.0 & 34.9 & 32.2 & 30.1 & 61.1 & 63.0 & 58.9 \\
\textbf{\method} & 42.5 & 50.0 & \textbf{47.5} & 45.8 & 40.3 & \textbf{36.7} & 55.7 & 62.3 & \textbf{70.0} \\
% END MAIN RESULTS PANEL B
\bottomrule
\end{tabular}
\end{table*}

\begin{table*}[t]
\centering
\caption{Self-evolving skill methods as instances of one train--select--test loop.
    Native update and retention rules are preserved; under our protocol every final artifact is additionally selected on validation and frozen before test.}
\label{tab:method-framework}
\small
\setlength{\tabcolsep}{4pt}
\begin{tabularx}{\textwidth}{@{}l>{\raggedright\arraybackslash}p{0.21\textwidth}
>{\raggedright\arraybackslash}X>{\raggedright\arraybackslash}p{0.25\textwidth}@{}}
\toprule
Method & Persistent artifact & Update from training feedback & Native candidate retention \\
\midrule
SkillGen & one skill & contrast successful and failed trajectories; iterative refinement & paired repairs minus regressions; deployment gate \\
Trace2Skill & skill directory & parallel trajectory-local patches; hierarchical consolidation & deduplication and conflict resolution \\
SkillOpt & one compact document & minibatch reflection; bounded add/delete/replace edits & strict validation; rejected-edit memory \\
EvoSkill & repository of skill folders & failure diagnosis; create or edit one skill & fixed-capacity validation frontier \\
GEPA & any textual component & reflective mutation from trajectory feedback; optional merge & Pareto frontier of candidates \\
\midrule
\method (Sec.~\ref{sec:method}) & metaskill; transient task-local skill & first-fault attribution; single-module edit & regression-aware paired gate \\
\bottomrule
\end{tabularx}
\end{table*}

\subsection{Controlled Protocol}
\label{app:protocol}

Each method is evaluated through three stages.
Training tasks expose the feedback permitted by the method's native update rule.
Validation tasks may be used to select a checkpoint or trigger a native stopping decision.
The selected artifact is then frozen before evaluation on test tasks.
Test outcomes cannot modify the artifact, update rule, stopping decision, or hyperparameters.

\begin{table}[h]
\centering
\caption{Information boundary across the three evaluation stages.}
\label{tab:appendix-access-boundary}
\small
\begin{tabularx}{\columnwidth}{lXX}
\toprule
Stage & Information available & Output of the stage \\
\midrule
Training & Task prompt, permitted inputs, execution feedback, and method-native update state & Candidate skill artifacts and checkpoints \\
Selection & Validation tasks and their executable outcomes & Frozen artifact, native stop decision, and artifact identifier \\
Test & Test prompt, permitted inputs, and execution environment & Final task outcomes; no subsequent method update \\
\bottomrule
\end{tabularx}
\end{table}

The comparison uses one model for every call: Qwen3-Coder-480B-A35B-Instruct-FP8 with deterministic decoding.
Within each benchmark, methods use the same task partitions, executor, visible feedback, and task-level limits.
The full artifact is passed to the executor without silent truncation.
Methods retain their native update operators, checkpoint cadence, and stopping rules.
We match the execution interface, information boundary, and declared task-level caps.
Realized calls and artifact lengths are retained as accounting fields.

\subsection{Benchmarks and Splits}
\label{app:benchmarks}

The evaluation spans six benchmarks.
Every test partition is task-disjoint from training and validation.
Family-disjointness is a stronger condition used only where the benchmark protocol supports it.
RBioBench Clinical and RBioBench Omics are two benchmarks constructed by the authors and part of the same RBioBench suite.
They share the task format and verifier but cover different workflow families and are reported in separate columns.

\begin{table*}[t]
\centering
\caption{Expanded benchmark inventory.
    Counts denote training/validation/test tasks.}
\label{tab:appendix-benchmark-inventory}
\footnotesize
\setlength{\tabcolsep}{3pt}
\begin{tabularx}{\textwidth}{>{\raggedright\arraybackslash}p{0.18\textwidth}c>{\raggedright\arraybackslash}p{0.18\textwidth}X}
\toprule
Domain & Tasks & Held-out boundary & Output and evaluation \\
\midrule
SpreadsheetBench & 40/20/40 & Task; shared operation types & Edited workbook; workbook content and structure checks \\
RBioBench & 40/20/40 & Task; shared package families & Scientific workflow outputs; RBioBench evaluator for Clinical and Omics tracks \\
SWE-bench Verified & 40/20/40 & Test repository & Source-code patch; repository-specific test suite \\
HealthBench & 40/20/40 & Medical specialty & Retrieved evidence and response; family-balanced rubric score plus artifact gate \\
SciVisAgentBench & 66/21/20 & Data family & Visualization artifact; family-balanced artifact evaluator \\
\bottomrule
\end{tabularx}
\end{table*}

SpreadsheetBench samples distinct task IDs from separate official development and test pools, while cell- and sheet-level operation types recur across its partitions.
RBioBench uses distinct task IDs selected by deterministic, outcome-blind stratification over track and task level; package families may recur across training, validation, and test.
SWE-bench uses four repositories for training and validation and two different repositories for test.
HealthBench assigns disjoint medical specialties to all three partitions.
SciVisAgentBench assigns each of its 84 eligible data families to exactly one partition.

SpreadsheetBench requires agents to inspect and modify workbooks while preserving relevant structure.
RBioBench requires executable clinical and omics workflows whose output files are checked by the RBioBench evaluator.
SWE-bench Verified evaluates repository repair through repository-specific tests.
SciVisAgentBench evaluates generated scientific visualizations and their artifacts.
Section~\ref{app:healthbench} specifies our HealthBench setup, including its specialty-disjoint split, retrieval contract, artifact checks, and family-balanced physician-rubric score.

\paragraph{RBioBench Clinical and Omics.} RBioBench is a benchmark we built to test whether agents can write working R programs for real clinical and bioinformatics analyses.
The version used in this paper has 405 tasks in two tracks.
The Clinical track (197 tasks) uses 12 packages from the pharmaverse ecosystem for clinical-trial data, led by \texttt{admiral} (113 tasks) and \texttt{aNCA} (52).
The Omics track (208 tasks) uses 38 CRAN and Bioconductor packages, led by \texttt{maftools} (30), TCGA data utilities (18), and \texttt{Biostrings} (11).
Each task was written around one real package function or workflow and has four parts: a prompt that names the inputs and the required output files, input files, a reference R solution, and the expected outputs produced by that solution.
Tasks come in three levels, from a single function call (L1, 234 tasks) to multi-step workflows (L2, 136; L3, 35).
Outputs are checked by exact comparison for 356 tasks, by a tolerance-based comparator for 30, and by a semantic check for 19.
For this study we treat the two tracks as two benchmarks, Clinical and Omics.
The study forms fixed training, validation, and test partitions by deterministic, outcome-blind stratification over benchmark track and task level.
Eligibility requires a declared input inventory, nonempty expected artifacts, and a passed prompt--artifact contract audit.
Tasks with prompt/header mismatches or missing previewed inputs are excluded before method evaluation.

Both benchmarks use the same mini-swe-agent v2 execution interface, a minimal agent from the SWE-agent team~\citep{yang2024sweagent}, a network-disabled pinned R/Bioconductor container, and the RBioBench task verifier.
A task passes when the required output artifacts are produced and accepted by its task-defined checks; missing outputs, execution errors, and failed artifact comparisons remain failures.
Before each run, we check manifest identity, task membership, fixtures, and evaluator inputs, while per-task workspaces isolate generated files during execution.
The experiment fixes the RBioBench source version and records runner, evaluator, manifest, and container identities in the execution ledger.

\subsection{GSO Procedure}
\label{app:gso-procedure}

The persistent object in \method is a metaskill, whereas the artifact executed on a task is transient.
For each visible task, the procedure is:

\begin{enumerate}
    \item analyze the task objective, input inventory, environment, and output contract;
    \item identify applicable tools and source evidence available in the task environment;
    \item compile one task-local executable skill from the frozen metaskill;
    \item execute the compiled skill with the common solver and executor;
    \item validate the produced artifact against task-visible requirements;
    \item if needed, perform at most one bounded recovery within the episode; and
    \item discard the task-local skill after recording the execution outcome.
\end{enumerate}

Task-local construction may use the prompt, provided files, observable schema, installed interfaces, and local documentation.
It may not inspect gold artifacts, reference solutions, hidden cases, or evaluator implementation.
When available evidence does not justify a binding, the task-local skill must preserve the uncertainty or abstain rather than convert a guess into a persistent global rule.

Persistent updates are proposed only after failure attribution.
Let $m$ be the parent metaskill and $m'$ a candidate that changes one attributed module.
On paired validation tasks, a repair is a task on which $m$ fails and $m'$ passes; a regression is a task on which $m$ passes and $m'$ fails.
The promotion utility is
\begin{equation}
u(m',m)=|R|-2|B|,
\end{equation}
where $R$ and $B$ are the repair and regression sets.
A candidate is promoted only when $u>0$, total validation success does not decrease, and the edited module matches the attributed cause.
Otherwise the parent is retained.

\subsubsection{Continuous-score gates}

For SciVisAgentBench, let $\delta_i$ be the paired candidate-minus-parent score on validation task $i$.
We use the magnitude-aware utility
\begin{equation}
u_{\mathrm{cont}}(m',m)=\sum_i\max(\delta_i,0)
-2\sum_i\max(-\delta_i,0),
\end{equation}
and require the family-balanced validation mean to increase.
HealthBench instead requires a positive paired family-balanced score change, a positive mean change in at least three of five validation specialties, no specialty mean below $-0.10$, and an additional penalty for safety-critical regression.
These gates keep the same regression-aware rule while respecting each benchmark's native outcome.

\subsubsection{Initialization and persistence scope}

Each benchmark-specific \method trajectory starts from the same declared, human-authored initial metaskill $m^{(0)}$.
Accepted updates persist within that trajectory and produce the selected $m^*$; they are not carried from one benchmark domain into another.
The reported \method condition executes task-local skills compiled from this selected artifact.
The complete $m^{(0)}$ is reproduced below, with its wording preserved and Markdown structure reformatted for LaTeX.

\paragraph{Task Classification.} \textbf{Inputs:} visible objective, declared inputs, available tool/environment interfaces, required artifact or state transition, and completion criteria.
\textbf{State:} \texttt{objective}, \texttt{inputs}, \texttt{interfaces}, \texttt{required\_outputs}, \texttt{completion\_checks}, and \texttt{unknowns}.
\textbf{Policy:} classify the task before selecting a procedure.
Keep uncertain fields explicit rather than guessing them, and distinguish task evidence from merely available context.
\textbf{Output:} one normalized task profile consumed by the remaining modules.

\paragraph{Source and Tool Applicability.} \textbf{Inputs:} normalized task profile, task-visible local evidence, installed interfaces/help, and authoritative documentation available to the executor.
\textbf{State:} an evidence ledger containing each consequential binding, its source class, support status, applicability conditions, conflicts, and unresolved fields.
\textbf{Policy:} identify which evidence supports each consequential tool, API, schema, and data binding before selecting a procedure.
Prefer stable public interfaces.
Treat merely available files, tools, or examples as inspect-only until task evidence establishes their role.
\textbf{Output:} an applicability-ranked source and tool plan with unresolved bindings kept explicit.

\paragraph{Transient Task-Specific Skill Construction.} \textbf{Inputs:} normalized task profile and applicability-ranked evidence plan.
\textbf{State:} \texttt{procedure}, evidence-backed local bindings, visible output contract, execution plan, validation plan, negative scope, and termination checks.
\textbf{Policy:} construct exactly one transient task-specific skill from this metaskill and the visible task.
Select the shortest procedure whose consequential bindings are supported; retain unknowns as runtime checks.
\textbf{Output:} one complete task skill used once for the current task.
It is never revised or promoted into persistent memory.

\paragraph{Executable Validation.} \textbf{Inputs:} one transient task skill and the task-visible executor state.
\textbf{State:} current step, tool observations, artifact/state evidence, declared checks, and terminal status.
\textbf{Policy:} execute the selected procedure and validate the required artifact or state using task-visible checks.
Status narration is not completion: continue tool use, including polling long-running commands, until execution and declared validation reach a terminal result.
\textbf{Output:} a verified terminal artifact/state, or a structured first-fault record; prose claiming completion is not an output.

\paragraph{Failure Attribution.} \textbf{Inputs:} task profile, task skill, ordered actions/observations, artifact state, and task-visible validation result.
\textbf{State:} \texttt{target\_stage}, \texttt{failure\_mechanism}, supporting observations, evidence strength, applicability scope, preservation constraints, and confounder status.
\textbf{Policy:} preserve the first actionable fault and attribute it to contract parsing, source selection, task-skill compilation, execution, artifact validation, or bounded recovery.
Do not replace an earlier actionable cause with the later symptom of a missing final artifact.
\textbf{Output:} one normalized learning signal.
Unsupported or infrastructure-only failures are marked non-learnable rather than converted into policy.

\paragraph{Bounded Repair and Abstention.} \textbf{Inputs:} normalized learning signal, current task state, and the behavior that must be preserved.
\textbf{State:} one repair hypothesis, its target stage, negative scope, repair count, and expected observable.
\textbf{Policy:} attempt at most one bounded repair that targets the attributed fault while preserving unrelated successful behavior, then rerun the declared validation.
If evidence remains insufficient or conflicting, abstain with concrete blocking evidence.
\textbf{Output:} a validated repaired task state or an evidence-backed abstention.
Never encode task identifiers, benchmark identities, hidden evaluator behavior, literal answers, or one-task exceptions as reusable policy.

\subsection{Metrics and Accounting}
\label{app:accounting}

For an artifact $s$ and split $q$, $a_q(s)$ denotes the fraction of tasks that pass the benchmark's executable verifier.
Main-table percentages preserve the denominator defined for each benchmark and split.
Figure~\ref{fig:train-test-comparison} plots the descriptive difference
\begin{equation}
D_{\mathrm{train}\rightarrow\mathrm{test}}(s)
=100\,[a_{\mathrm{test}}(s)-a_{\mathrm{train}}(s)].
\end{equation}
This quantity directly compares the two reported scores and is interpreted alongside the checkpoint trajectories.

Training gain, test gain, and retention follow Section~\ref{sec:generalization}, with $s_0$ the empty skill, so $a_q(s_0)$ is the No Skill row of Table~\ref{tab:main-results}.
Retention is therefore each method's $D_{\mathrm{train}\rightarrow\mathrm{test}}$ minus that of No Skill on the same benchmark.
Across the six test splits, 21 of the 36 baseline artifacts gain on their training tasks; 5 have retention at or above zero, 13 have negative retention with a positive test gain, and 3 have no test gain: RBioBench Omics EvoSkill ends level with No Skill, and SpreadsheetBench GEPA and SWE-bench Trace2Skill end below it.
Retention ranges from $+3.4$ (RBioBench Clinical EvoSkill) to $-15.0$ (SpreadsheetBench EvoSkill).

For paired artifact comparisons, every task belongs to exactly one of four states: repair, regression, both pass, or both fail.
Reporting only aggregate accuracy can hide this distinction; two artifacts with the same pass count can repair and break different tasks.
We therefore retain paired states whenever the same task set is evaluated under both artifacts.
Infrastructure or artifact contract failures remain failures under the declared protocol and are not silently removed from the denominator.

Binary domains define repairs and regressions from verifier outcomes.
For SciVisAgentBench, $\delta_i$ denotes the paired candidate-minus-parent score on task $i$, and the promotion utility is
\begin{equation}
u_{\mathrm{cont}}=\sum_i\max(\delta_i,0)
-2\sum_i\max(-\delta_i,0).
\end{equation}
Promotion also requires an increase in the family-balanced validation mean.
HealthBench instead requires a positive paired family-balanced score change, positive mean change in at least three of five validation specialties, no specialty mean below $-0.10$, and an additional penalty for safety-critical regression.

Task-level accounting binds each aggregate entry to the benchmark, split, method, denominator, seed, executed artifact, and verifier outcome.
Checkpoint analyses and final comparisons retain distinct evidence roles throughout the evaluation.

\subsection{Judge Protocol}
\label{app:judge}

The judge reads only the skill, independently of execution outcomes and \method's update decision.

Table~\ref{tab:judge-rubric} in the main text lists the five dimensions, their weights, and the question each one asks.

Each dimension receives a score from 1 to 5, an artifact excerpt, and a one-sentence rationale.
The prompt instructs the Judge not to reward length or surface fluency.
Generalizability is capped when task identifiers, literal answers, fixed layouts, or evaluator-specific tricks are presented as general rules.

Figure~\ref{fig:checkpoint-judge-generalization} contains 126 predefined final-artifact pairs.
For each pair, $A$ is the artifact with the higher pointwise Judge score and $B$ is the other artifact.
The axes are
\begin{align}
x &= 100\,[a_{\mathrm{test}}(A)-a_{\mathrm{test}}(B)],\\
y &= J(A)-J(B).
\end{align}
Thus, points with $x>0$ indicate agreement between the Judge-induced ordering and executable test performance.
These labels are induced from pointwise scores; they are not direct pairwise Judge calls.
No Skill is excluded because it has no skill artifact to score.
The pairwise comparisons reuse the same scored artifacts within each domain.
Across the six domains, the Judge-induced ordering agrees with test performance for 108/126 pairs (85.7\%).

A separate checkpoint analysis evaluates 45 parent-to-candidate transitions: 21 from SpreadsheetBench and 24 from RBioBench.
Judge utility is defined as $u_J=p_J(\mathrm{repair})-2p_J(\mathrm{regression})$, and the observed test change is $100[a_{\mathrm{test}}(\mathrm{candidate})- a_{\mathrm{test}}(\mathrm{parent})]$.
The corresponding Spearman correlations are 0.39 on SpreadsheetBench and approximately zero on RBioBench.
We therefore use this analysis to diagnose artifact content, while executable verifiers determine task correctness and method selection.

\subsection{HealthBench Retrieval}
\label{app:healthbench}

The HealthBench experiment evaluates a separately named retrieval workflow over specialty-disjoint training, validation, and test sets.
For each conversation, the agent may search for evidence, select supporting sources, construct a response, and submit the required retrieval artifacts.
The same retrieval and artifact contract is applied to all methods, and every submitted response remains in the declared denominator.

The reported score is the family-balanced physician-rubric score.
On the 40 test conversations, \method raises the score from 23.94\% (No Skill) to 36.72\%.
All five test specialties have a positive mean difference under \method.

\subsection{Cases and Reproducibility}
\label{app:cases}

Figure~\ref{fig:skill-content} in the main text separates three evidence layers.
Text inside the artifact cards is a verbatim excerpt.
Labels describing placeholders, global checklists, or feedback-specific rules are author interpretations of those excerpts.
PASS labels report the observed outcome of the displayed task-local executions.
The cases show how visible task evidence instantiates concrete objects, operations, parameters, and read-back checks.

The Biostrings example binds the semantic transformation \texttt{DNAStringSet}$\rightarrow$\texttt{AAStringSet}, executes translation, writes the required object, and reloads it for validation.
The ShortRead example binds a low-quality-end trimming operation, its parameters, the read/write sequence, and a post-write quality check.
In both cases, the persistent artifact supplies construction policy while the transient skill contains the task-specific scientific bindings.

\paragraph{Figure inventory.}
\label{app:reproducibility}

All quantitative figures are generated with Python/matplotlib.
Vector exports retain editable text, and each accepted figure is accompanied by the plotted rows and a machine-readable generation record.
Each figure has one evidence role, listed in the table below.

\begin{table*}[t]
\centering
\caption{Figure-level evidence and interpretation boundaries.}
\label{tab:appendix-figure-inventory}
\small
\begin{tabularx}{\textwidth}{clXX}
\toprule
Figure & Primary input & Evidence role & Interpretation boundary \\
\midrule
\ref{fig:story} & Two candidate edits from one SkillOpt run & Illustrates similar training gains with opposite test effects & Two edits from one run; not a prevalence estimate \\
\ref{fig:train-test-generalization-case} & Stored optimization checkpoints & Shows development and test trajectories & Descriptive post-freeze analysis; test does not select artifacts \\
\ref{fig:train-test-comparison} & Main result table train/test columns & Shows signed train-to-test score differences & Reports the two displayed split scores directly \\
\ref{fig:skill-content} & Artifact excerpts and execution traces & Illustrates task-local binding and validation & Qualitative execution evidence from the displayed cases \\
\ref{fig:gso-framework} & Method description & Shows the \method loop & Schematic; no data \\
\ref{fig:domain-grouped-main-results} & Main result table & Compares final test performance across domains & Preserves benchmark-specific metrics and denominators \\
\ref{fig:reasonable-judge} & Frozen artifact texts and pointwise rubric & Compares artifact-content properties & Does not measure executable correctness \\
\ref{fig:checkpoint-judge-generalization} & Predefined artifact pairs and test scores & Tests Judge-induced ordering against outcomes & Retrospective association, not a selection rule \\
\bottomrule
\end{tabularx}
\end{table*}

For quantitative results, source rows retain method names, split labels, reported values, and the mapping to the displayed panel.
Figure exports are produced as PDF and SVG for manuscript use and as high-resolution PNG and TIFF for visual inspection.
This separation between source rows, generation code, and manuscript assets makes numerical changes detectable while keeping the paper focused on reproducible, task-level evidence.

\section{Extended Related Work}
\label{app:related-work}

This section expands Section~\ref{sec:related-work}.
We group prior work by two questions: what object persists across tasks, and when that object is allowed to change.
Together these decide what "generalization" can mean for a given system, which is why the main text fixes both before measuring it.

\paragraph{Agent memory and retained experience.} The earliest systems in this line keep a store of past episodes and reflect on it.
Generative agents store experiences in a memory stream, turn them into reflections, and retrieve them to plan \citep{park2023generative}.
Reflexion writes verbal feedback after a failed attempt and reads it on the next one; ExpeL extracts natural-language insights from training tasks; MetaReflection turns past reflections into reusable instructions \citep{shinn2023reflexion,zhao2024expel,gupta2024metareflection}.
Contextual Experience Replay turns past experience into a memory of environment dynamics and common decision patterns for later retrieval \citep{liu2025cer}, and ReasoningBank distills reasoning strategies from both successes and failures \citep{ouyang2026reasoningbank}.
A more recent group learns how to manage the memory itself: Memory-R1 trains explicit memory operations with reinforcement learning, MetaMem evolves a meta-memory that carries experience in using knowledge across tasks, and the meta-cognitive memory copilot of \citet{liang2026remember} studies which abstraction level transfers across tasks and when transfer is negative \citep{yan2026memoryr1,xin2026metamem}.
Experience can also be stored in weights as composable latent modules \citep{ling2026reusable}.
What these systems share is the assumption that more retained experience helps later tasks; the memory-management work already shows that this is not automatic.
We fix the retained object to a skill, hold everything else constant, and measure how much of a development gain reaches frozen held-out tasks.

\paragraph{Reusable skills, tools, and workflows.} A second line stores procedures rather than notes.
Skill induction with latent language learns reusable skills described in natural language \citep{sharma2022skill}; Voyager grows an executable code library through play in an open world; Agent Workflow Memory induces reusable workflows from web trajectories \citep{wang2023voyager,wang2024awm}.
LATM writes a tool once and reuses it on later instances \citep{cai2024latm}, CREATOR separates the abstract act of making a tool from the concrete act of running it \citep{qian2023creator}, and ToolMaker turns published code into agent tools with closed-loop repair \citep{wolflein2025toolmaker}.
Memp, MUSE, Mem$^2$Evolve, and AgentFactory keep procedural or experience memory, and in some cases tools or subagents, that is refined as tasks accumulate \citep{fang2026memp,yang2026muse,cheng2026mem2evolve,zhang2026agentfactory}, and SAGE uses reinforcement learning over chains of similar tasks to train the model to generate and use skills from a growing library \citep{wang2026sage}.
The four skill optimizers we evaluate, SkillGen, Trace2Skill, SkillOpt, and EvoSkill, all revise an explicit skill from execution feedback \citep{ma2026skillgen,ni2026trace2skill,yang2026skillopt,alzubi2026evoskill}; Table~\ref{tab:method-framework} lists their update and retention rules.
The separation that tool-making already makes, between a reusable maker and a disposable instance, is the same separation \method makes between a metaskill and a transient task skill.
The difference is that \method keeps the reusable object as a guide for writing procedures and never promotes a task skill into it.

\paragraph{Prompt, program, and agent-design optimization.} If a skill is read by a model before it acts, then prompt optimizers are the obvious comparison.
OPRO and ProTeGi optimize prompts from a history of candidates and from textual gradients; EvoPrompt applies evolutionary search \citep{yang2023opro,pryzant2023protegi,guo2024evoprompt}.
DSPy and MIPRO treat a pipeline as a program whose instructions and demonstrations are compiled against a metric, and PROMST does the same for long prompts in multi-step agent tasks \citep{khattab2024dspy,opsahlong2024mipro,chen2024promst}.
TextGrad passes textual feedback through compound systems, and GEPA reflects on trajectories and keeps a Pareto frontier of candidate prompts \citep{yuksekgonul2025textgrad,agrawal2025gepa}.
One level up, AFlow and AgentSquare search over workflows and modular agent components, ReCreate edits domain-agent scaffolds from execution experience, and textual backpropagation builds multi-agent teams layer by layer and refines each agent's role, prompt, and coordination \citep{zhang2025aflow,shang2025agentsquare,hao2026recreate,ma2026textualbackprop}.
These optimizers report the score their objective reaches on the tasks they optimized on, or on a held-out split of the same distribution, and they are usually careful about it.
What they do not report is how the score moved during optimization on tasks the optimizer could not see.
We track that trajectory for every method, and we include GEPA as the optimizer baseline because it is the strongest general-purpose one in this group.

\paragraph{Test-time learning and self-improvement.} Several systems keep learning after deployment.
Dynamic Cheatsheet maintains an evolving memory at inference time, and test-time-learning evaluations compare limited with cumulative experience \citep{suzgun2026dynamic,wang2025experience}.
Human-in-the-loop guidance asks human experts for corrections at test time \citep{he2025hitl}, and CURE trains one model to critique its own answer and retry \citep{chen2026cure}; RISE trains models to improve over repeated attempts \citep{qu2024rise}; Search-R1 and DeepRetrieval train policies that treat retrieval as an environment \citep{jin2025searchr1,jiang2025deepretrieval}; SkillTTA synthesizes a temporary task-conditioned skill at test time from retrieved training trajectories \citep{wang2026skilltta}.
Recursive self-improvement systems modify their own scaffolds, programs, or logic \citep{zelikman2024stop,yin2025godel,zhang2025dgm}, and ADAS uses a meta agent that programs new agents in code \citep{hu2024adas}.
In all of these, a held-out score mixes two effects: what the artifact learned before deployment and what it learned during it.
Our protocol freezes the artifact before test so that only the first effect is measured.
SkillTTA is the closest in spirit to \method's transient skill; the difference is that \method writes the task skill from a learned guide and the task's own evidence, without retrieving training trajectories at test time.

\paragraph{Evaluating experience transfer.} Negative results about transfer are beginning to appear.
Experience-following studies show that retained memories propagate errors and that replaying experience on tasks it does not match gives limited or even misleading value \citep{xiong2026memorymanagement}.
TraineeBench evaluates learning, exploration, and scheduling in dynamic workplace scenarios, and TRACE evolves benchmark tasks that are validated by reproducible trajectories \citep{fu2026agentfirstday,guo2026trace}.
LLM judges allow qualitative comparison at scale but are sensitive to presentation order \citep{zheng2023mtbench,shi2024judging}.
Our evaluation adds two things to this work: checkpointed development scores paired with frozen held-out execution for several skill-evolution methods at once, and a content judge that is used only after the fact, as a diagnostic, and never as an outcome or a selection signal.

\end{document}